\documentclass{article}
\PassOptionsToPackage{square, numbers}{natbib}
\usepackage[main,final]{neurips_2026}
\usepackage[square, numbers]{natbib}
\usepackage[utf8]{inputenc} % allow utf-8 input
\usepackage[T1]{fontenc}    % use 8-bit T1 fonts
\usepackage{hyperref}       % hyperlinks
\usepackage{url}            % simple URL typesetting
\usepackage{booktabs}       % professional-quality tables
\usepackage{amsfonts}       % blackboard math symbols
\usepackage{nicefrac}       % compact symbols for 1/2, etc.
\usepackage{microtype}      % microtypography
\usepackage{xcolor}         % colors
\usepackage{tcolorbox}
\usepackage{multirow}
\usepackage{makecell} 
\usepackage{enumitem}
\usepackage{titletoc}
 \usepackage{amsmath}
\title{Benchmarking Vietnamese Legal Knowledge of Large Language Models}

\author{
  \textbf{Nguyen Tien Dong}\textsuperscript{1,2,*} \quad
  \textbf{Minh-Anh Nguyen}\textsuperscript{1,2,*} \quad
  \textbf{Thanh Dat Hoang}\textsuperscript{1} \\
  \textbf{Nguyen Tuan Ngoc}\textsuperscript{1} \quad
  \textbf{Dao Xuan Quang Minh}\textsuperscript{1} \quad
  \textbf{Phan Phi Hai}\textsuperscript{1} \\
  \textbf{Nguyen Thi Ngoc Anh}\textsuperscript{1,3,\dag} \quad
  \textbf{Binh Vu}\textsuperscript{4,\dag} \quad
  \textbf{Dung D. Le}\textsuperscript{2,\dag} \\[0.6em]
  \textsuperscript{1}CMC OpenAI \quad
  \textsuperscript{2}VinUniversity \quad
  \textsuperscript{3}HUST, Vietnam \\
  \textsuperscript{4}SRH University Heidelberg, Germany \\[0.4em]
  \texttt{\{dongnt,minhna\}@cmcai.vn}, \;
  \texttt{\{25dong.nt,minh.na2\}@vinuni.edu.vn} \\[0.3em]
  {\small $^*$Co-first authors. \quad $^\dag$Co-last authors.}
}

\begin{document}
\maketitle
\begin{abstract}
The rapid advancement of large language models (LLMs) has expanded their potential in the legal domain. However, existing legal benchmarks remain largely English-centric and oriented toward common law, leaving a critical gap in evaluating LLMs for civil law systems that govern most jurisdictions worldwide. To address this gap, we introduce Vietnamese Legal Benchmark (\textbf{VietLegal}), a cognitively grounded benchmark designed for the hierarchical and codified structure of Vietnamese law. Although instantiated in Vietnamese legislation, VietLegal provides a replicable evaluation framework for civil law systems characterized by complex statutory hierarchies and frequent amendments. Inspired by Bloom’s taxonomy, VietLegal assesses multiple levels of legal understanding through tasks that mirror real-world legal assistant use cases, including legal question answering, multi-step reasoning, and scenario-based problem solving. The benchmark contains 10,450 expert-annotated samples, each cross-validated against authoritative legal sources to ensure fidelity to practical legal workflows. By offering the first standardized legal benchmark for Vietnamese, VietLegal enables systematic assessment of LLMs in civil law contexts and supports the development of more reliable and interpretable AI-assisted legal systems.
\end{abstract}
\section{Introduction}

% \begin{figure*}[ht]
%     \centering
% \includegraphics[width=0.7\textwidth]{img/VietLegalBench_overview.png}
%     \caption{The five-level cognitive framework of VietLegal.}
%     \label{fig:overview_bench}
% \end{figure*}
The rapid progress of large language models (LLMs) has enabled transformative applications in the legal domain \cite{homoki2024large, sun2023short}. While LLMs demonstrate strong performance on general tasks, their effectiveness in legally complex and low-resource languages like Vietnamese remains largely unexplored. Vietnamese law is characterized by a formal, hierarchical, and continuously evolving statutory system, requiring specialized evaluation to ensure that model 
outputs remain legally accurate, consistent, and ethically aligned. 

Existing legal NLP benchmarks predominantly target common law in English, emphasizing case-based reasoning \cite{chalkidis2022, guha2023}. This focus overlooks civil law systems, which govern over 60\% of global jurisdictions and derive authority from hierarchical statutes rather than judicial precedent \cite{merryman2018civil}. Civil law introduces distinct challenges, requiring models to navigate complex statutory interpretations and track temporal validity across frequent amendments. While recent benchmarks have addressed Chinese civil law \cite{li2024lexeval, dai2025laiw, fei2024lawbench,li2025}, other codified traditions remain underrepresented. Vietnamese law, specifically, presents unique difficulties due to its heavy reliance on intricate references among \textit{Articles}, \textit{Clauses}, and \textit{Points}, requiring specialized evaluation to ensure legal fidelity.

To address these limitations, we introduce \textbf{VietLegal}, the first comprehensive benchmark designed to evaluate LLMs on Vietnamese legal tasks within a civil law framework. Grounded in Bloom’s cognitive taxonomy, VietLegal assesses model capabilities across progressively deeper levels ranging from basic recall to multi-step reasoning and ethical judgment. The benchmark contains 10,450 expert-annotated samples, each cross-validated against authoritative legal sources to ensure fidelity to practical legal workflows. By offering the first standardized legal benchmark for Vietnamese, VietLegal enables systematic assessment of LLMs in civil law contexts and supports the development of more reliable AI-assisted legal systems.

Our main contributions are summarized as follows: First, we introduce VietLegal, a benchmark for evaluating LLMs on Vietnamese legal tasks with a replicable design applicable to other civil law jurisdictions. Second, we propose a cognitively grounded evaluation methodology informed by Bloom’s taxonomy that enables systematic assessment from basic legal recall to advanced multi-step reasoning. Third, we release a high-quality dataset of 10,450 expert-verified samples and conduct extensive experiments across 23 diverse LLMs, offering insights into their strengths and limitations in civil law reasoning. The benchmark and evaluation code are available at \url{https://huggingface.co/datasets/CMC-OPENAI/VLegal-Bench}.

\begin{table*}[!t]
\centering
\scriptsize
\setlength{\tabcolsep}{1.5pt}
\caption{Overview of VietLegal: The benchmark evaluates legal LLMs across five levels, from basic recognition to ethical reasoning, using five question templates: Multiple-Choice Question Answering (MCQ), Multi-Label Classification (MLC), Binary Classification (BC), Named Entity Recognition (NER) and Generation for Vietnamese law.}
\label{tab:benchmark_tasks}
\begin{tabular}{@{}p{2.3cm}cp{3cm}p{7cm}ccc@{}}
\toprule
\textbf{Level} & \textbf{ID} & \textbf{Task} & \textbf{Purpose} & \textbf{Type} & \textbf{Metric} & \textbf{Test set} \\
\midrule
\textbf{1. Recognition \& Recall} 
& 1.1 & Legal Entity Recognition & To detect and classify named entities, including persons, organizations, monetary amounts, ... within legal documents. & MCQ, NER & \makecell{Accuracy, \\ EM} & 750 \\
& 1.2 & Legal Topic Classification & Classifies legal questions into predefined legal topics. & MCQ, MLC & \makecell{Accuracy, \\Macro-F1} & 700 \\
& 1.3 & Legal Concept Recall & Recalls statutory definitions or meanings of legal terms and concepts. & MCQ & Accuracy & 300\\
& 1.4 & Article Recall & Retrieves or cites the correct legal article corresponding to a term, concept, or question. & MCQ & Accuracy & 1000\\
& 1.5 & Legal Schema Recall & Recognizes and recalls hierarchical and temporal relations among legal documents (e.g., amendments, replacements, ...) & MCQ & Accuracy & 800\\
\midrule
\textbf{2. Understanding \& Structuring} 
& 2.1 & Relation Extraction & Extracts the subject, object, and content of a legal relationship from a factual scenario & MCQ & Accuracy & 253 \\
& 2.2 & Legal Element Recognition & Identifies the hypothesis, disposition, and sanction components within a legal provision & MCQ & Accuracy & 300 \\
& 2.3 & Legal Graph Structuring & Convert legal documents into structured knowledge graphs representing entities, relations, and inter-article references. & \makecell{Generat.,\\ MLC} & \makecell{ROUGE-L,\\ Node-F1, \\Edge-F1} & 296 \\
& 2.4 & Judgment Verification & Evaluates whether a court's reasoning or statement is consistent with the factual and legal content of the actual judgment. & BC & Accuracy & 600 \\
& 2.5 & User Intent Understanding & Determines the underlying intent or query type of the user when interacting with a legal assistant. & MLC & macro-F1 & 1359\\
\midrule
\textbf{3. Reasoning \& Inference} 
& 3.1 & Article / Clause Prediction & Predict which legal article or clause applies to a given legal question or short query, instead of a lengthy factual scenario & MCQ & Accuracy & 600\\
& 3.2 & Legal Court Decision Prediction & Predicts the final court decision or judgment outcome from the factual and legal content of a real case. & MCQ & Accuracy & 600\\
& 3.3 & Multi-Article Reasoning & Perform multi-step reasoning by connecting several legal provisions or facts to derive a consistent conclusion. & MCQ & Accuracy & 292\\
& 3.4 & Conflict \& Consistency Detection & Identify contradictions or overlaps between different legal clauses or interpretations across statutes or contracts. & BC & Binary F1 & 161\\
& 3.5 & Penalty / Remedy Estimation & Estimates the appropriate legal penalty or remedy for a given factual situation. & MCQ & Accuracy & 358\\
\midrule
\textbf{4. Interpretation \& Generation} 
& 4.1 & Legal Document Summarization & Generate concise summaries of long legal texts (statutes, judgments, contracts) while preserving key information. & Generat. & \makecell{ROUGE-L,\\ BertScore} & 384\\
& 4.2 & Judicial Reasoning Generation & Produce structured reasoning paragraphs based on the IRAC template (Issue - Rule - Application - Conclusion). & Generat. & \makecell{ROUGE-L,\\ BertScore} & 299\\
& 4.3 & Objective Legal Opinion Generation & Generate a balanced and impartial legal opinion or advisory text that aligns with statutory interpretation. & Generat. & \makecell{ROUGE-L,\\ BertScore} & 498\\
\midrule
\textbf{5. Ethics, Fairness \& Bias} 
& 5.1 & Bias Detection & Detect gender, racial, political, or religious bias in generated answers or decisions to ensure fairness. & MCQ & Accuracy & 250\\
& 5.2 & Privacy \& Data Protection & Identify and redact sensitive or personal data in legal texts to ensure privacy compliance. & MCQ & Accuracy & 216\\
& 5.3 & Ethical Consistency Assessment & Evaluate whether the model's outputs align with professional ethics and moral standards in legal reasoning. & MCQ & Accuracy & 200\\
& 5.4 & Unfair Contract Detection & Compare model judgments across similar cases or parties to assess impartiality and equitable reasoning. & MCQ & Accuracy & 234\\
\bottomrule
\end{tabular}
\end{table*}

\section{VietLegal}

\subsection{Design Principle of VietLegal}

\textbf{VietLegal} is organized around a hierarchical cognitive framework inspired by Bloom’s taxonomy and adapted to the linguistic and structural properties of Vietnamese law, comprising five levels of legal cognition ranging from factual recognition to advanced legal reasoning. Each task is explicitly designed to reflect challenges inherent to civil-law systems, with Levels~3 and~4 in particular targeting complex statutory reasoning phenomena such as multi-article dependency, hierarchical interpretation across legal instruments, and consistency analysis under overlapping or amended regulations. 

Although developed in Vietnamese, VietLegal provides a replicable framework for evaluating AI in codified legal systems. Its task design reflects realistic legal assistant use cases and targets core civil-law reasoning patterns rather than case-based analysis, enabling straightforward adaptation to other civil-law languages and jurisdictions. An overview of the benchmark is shown in Table~\ref{tab:benchmark_tasks}, with detailed task descriptions in Appendix~\ref{appen:task instruction}.

\begin{figure*}[t]
    \centering
\includegraphics[width=0.8\textwidth]{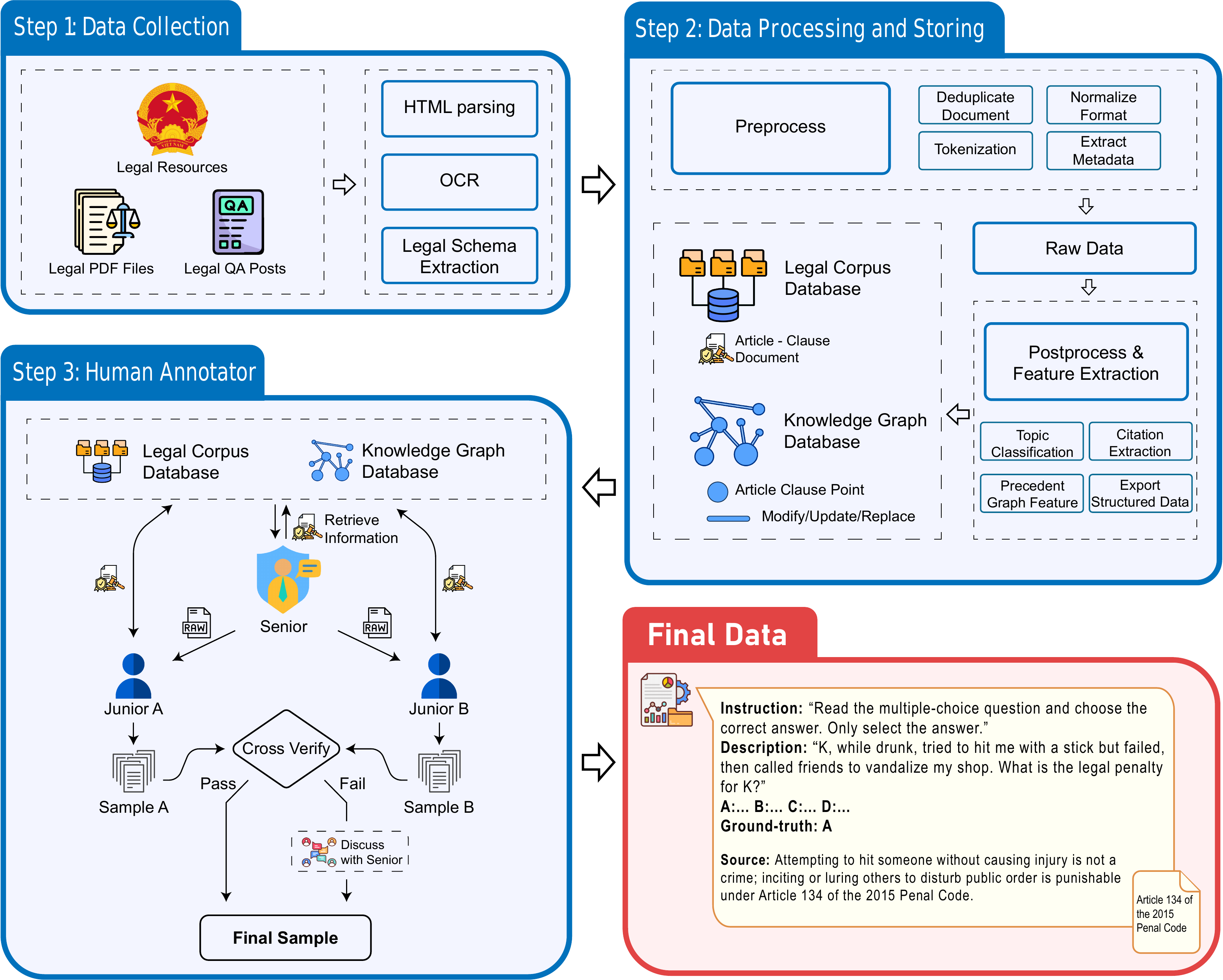}
    \caption{\textbf{VietLegal data pipeline.} Data are collected from Vietnamese legal sources, preprocessed, stored in a database, and used to build an information retrieval tool for legal experts. The final dataset is obtained through manual annotation.}
    \label{fig:data_process}
\end{figure*}

\textbf{Level 1 - Recognition \& Recall} targets foundational legal literacy in the Vietnamese context. It evaluates whether an LLM can accurately identify and retrieve core legal entities, concepts, and statutory provisions within dense and highly cross-referenced legal texts. These tasks assess basic factual competence, which is a prerequisite for deeper legal understanding, and simulate real-world interactions where users seek clarification of fundamental legal information.

\textbf{Level 2 - Understanding \& Structuring} examines an LLM’s ability to comprehend and organize complex statutory content. Given the hierarchical structure of Vietnamese law and its frequent amendments, this level evaluates whether models can capture relationships among \textit{Articles}, \textit{Clauses}, and \textit{Points}, and represent legal norms as a coherent, evolving system. The tasks reflect practical legal assistant scenarios, including analyzing long legal documents, verifying judicial decisions, and explaining statutory relationships to users.

\textbf{Level 3 - Reasoning \& Inference} assesses the model’s capacity to apply legal provisions to factual scenarios through logical and multi-step reasoning. Tasks at this level require predicting relevant articles, estimating penalties or remedies, synthesizing information across multiple statutes, and resolving conflicts between overlapping or amended legal norms. These skills are essential for realistic legal problem-solving and judicial support.

\textbf{Level 4 - Interpretation \& Generation} evaluates higher-order interpretive and generative abilities. This level tests whether an LLM can produce coherent, accurate, and unbiased legal texts, such as statute summaries, judicial analyses, and reasoned legal opinions. The tasks simulate professional legal workflows, including legal drafting, summarization, and scenario-based advisory reasoning using structured frameworks such as IRAC.

\textbf{Level 5 - Ethics, Fairness \& Bias} focuses on the normative dimensions of legal AI. It evaluates whether model outputs adhere to principles of fairness, impartiality, and privacy protection, particularly in sensitive or high-stakes legal contexts. These tasks stress-test an LLM’s ability to handle ethical dilemmas, bias-sensitive scenarios, and potentially unfair contractual or legal conditions, ensuring alignment with professional legal standards.

To further support construct validity, VietLegal is explicitly designed to reduce the effectiveness of superficial answer-selection strategies. All MCQ distractors are manually constructed by legal experts to be legally plausible, lexically similar to correct answers, and carefully targeted to reflect common reasoning errors rather than obvious misconceptions. In addition, we evaluate several non-reasoning baselines, including random guessing, a longest-answer heuristic, and semantic similarity using PhoBERT \cite{nguyen2020}. All of these approaches perform at or near chance levels across difficulty tiers, with some heuristics falling below chance due to adversarial distractor design. These findings indicate that surface-level cues are insufficient for solving the benchmark and that strong performance instead depends on genuine legal reasoning. Full experimental details are provided in Appendix~\ref{appen:construct_validity}.

% \paragraph{Task-to-level assignment.}
% The five levels are not abstract categories imposed on the data; they correspond to successive cognitive operations required to resolve a legal problem in a codified system: \emph{retrieving} the relevant legal information (L1), \emph{interpreting and organising} it (L2), \emph{applying} legal rules to facts (L3), \emph{articulating} the resulting conclusion as a complete legal analysis (L4), and \emph{evaluating} that analysis against professional, ethical, and fairness constraints (L5). Each task is assigned according to the \emph{dominant} cognitive operation required under its specified input and output conditions. For example, Judgment Verification (2.4), Court Decision Prediction (3.2), and Judicial Reasoning Generation (4.2) all involve a court case, yet they differ in what the model must produce: in 2.4 both facts and conclusion are given and the model only checks whether the conclusion is supported; in 3.2 the conclusion is withheld, so the model must identify and apply the governing rule; in 4.2 the model must additionally articulate the legal basis as a structured IRAC analysis, where a correct conclusion resting on a fabricated basis counts as incomplete. Importantly, the hierarchy orders tasks by the cognitive demand they place on \emph{human legal practitioners}, which is cumulative across levels; it is not a prediction of task difficulty for LLMs.
 
\paragraph{Independent expert validation.}
We developed the taxonomy iteratively in consultation with senior legal experts from two Vietnamese law firms. To assess its reproducibility, we recruited three additional senior legal experts who were not involved in benchmark design or annotation. Each expert independently assigned all 22 tasks to a taxonomy level using only the level definitions, with the original assignments withheld and task order randomised. Experts also rated the difficulty of each task for a competent human practitioner and their confidence in the assigned level on a 5-point scale. As reported in Table~\ref{tab:taxonomy_validation}, the independent assessments matched the original assignments in 64 of 66 cases (97.0\%), yielding a mean linear-weighted Cohen's $\kappa$ of 0.97. Human-difficulty ratings were also strongly correlated with the assigned taxonomy levels (Spearman's $\rho = 0.79$, $p < 10^{-4}$). The two disagreements arose for Judgment Verification (2.4) and Unfair Contract Detection (5.4), where one expert assigned L3 to each task. Both cases reflect ambiguity in determining whether a task requires rule application, suggesting that residual disagreement is concentrated at specific level boundaries rather than indicating a broader lack of reproducibility in the taxonomy.
 
\begin{table}[t]
\centering
\small
\caption{Independent expert validation of the task-to-level assignment (3 senior legal experts $\times$ 22 tasks = 66 judgments).}
\label{tab:taxonomy_validation}
\begin{tabular}{lc}
\toprule
\textbf{Measure} & \textbf{Value} \\
\midrule
Exact agreement with published level & 97.0\% (64/66; 95\% CI 92.4--100.0) \\
Agreement within $\pm$1 level & 98.5\% \\
Linear-weighted Cohen's $\kappa$ (mean over raters) & 0.97 (range 0.94--1.00) \\
Human difficulty vs.\ assigned level (Spearman $\rho$) & 0.79 ($p < 10^{-4}$) \\
Rater confidence in assignment (1--5) & 4.29 $\pm$ 0.92 \\
\bottomrule
\end{tabular}
\end{table}

\subsection{Data Collection and Processing}\label{sec:data_colection}

Legal documents were collected from official government portals and law firm Q\&A repositories, resulting in approximately 55{,}000 centrally issued and currently effective documents processed via HTML parsing and OCR (Figure~\ref{fig:data_process}). After deduplication and metadata extraction, we construct two complementary resources: (1) a \textbf{Knowledge Graph Database} that encodes the hierarchical structure of civil-law texts, including relationships among \textit{Articles}, \textit{Clauses}, and \textit{Points}, and (2) a \textbf{Legal Corpus Database} designed for scalable retrieval and analysis of long statutory documents. To ensure compliance with privacy and ethical standards, all personally identifiable information (PII) appearing in the source data is systematically anonymized; in particular, real names are replaced with standardized placeholders, e.g., "Le Thi X", following common conventions in Vietnamese legal and educational contexts. Together, these resources enable the systematic construction of tasks that directly address core civil-law challenges, such as hierarchical statutory interpretation, cross-article dependency, and consistency analysis under overlapping or amended regulations. This infrastructure supports 10{,}450 expert-annotated samples across 22 tasks, simulating realistic legal assistant workflows including statutory interpretation, legal drafting, and judicial decision support.

The labeling process followed a multi-stage protocol supervised by a senior lawyer and executed by independent junior legal experts. To maintain high fidelity, annotators performed blind cross-verification by exchanging batches every 100 samples, reaching an initial agreement of 92.39\% with a Cohen's Kappa of 0.89. Disputed cases (7.61\%) were resolved through structured junior consensus or escalated to senior adjudication, with unresolved cases concentrated in high-complexity tasks like conflict detection (Task 3.4) and multi-article reasoning (Task 3.3). The final benchmark comprises 10,450 expert-validated instances grounded in authoritative statutory sources, with additional details on the annotation support tool and annotator recruitment provided in Appendix~\ref{appen:annotate} and Appendix~\ref{appen:labelling}.

\section{Experiments and Results}\label{sec:full_experiment}
\subsection{Experiment Setting}
We evaluate large language models under both zero-shot and few-shot settings. In the zero-shot configuration, models receive only task instructions and the input query, while in the few-shot setting, we prepend a single task-specific demonstration example to each input instance. These demonstration examples are drawn from a separate development set and are not included in the reported test set. For both settings, we conduct evaluations with and without explicit reasoning: models are prompted either to produce a final answer directly or to generate intermediate reasoning using a chain-of-thought (CoT) prompt before outputting the final response. To ensure reproducibility, we fix the decoding temperature to 0, thereby minimizing variance introduced by stochastic sampling. The complete set of evaluation prompts is provided in Appendix~\ref{appen:task instruction}.

We adopt standardized evaluation metrics across tasks: Accuracy, F1 for multiple-choice and extraction-style questions, and ROUGE-L, BERTScore F1 using \textit{bert-base-multilingual-cased} for generation-level tasks. When the input length exceeds the maximum context window of an LLM, we apply middle truncation to the input sequence, as both the beginning and the end of legal texts often contain critical information. This truncation strategy follows prior legal LLM benchmarks~\cite{li2024lexeval,dai2025laiw,fei2024lawbench}.
\subsection{Evaluated Model}
We evaluate VietLegal on 23 LLMs spanning diverse model sizes, architectures, and training paradigms. The models are grouped into \emph{general multilingual LLMs} and \emph{Vietnamese-focused LLMs} based on their training focus. The multilingual group includes both proprietary closed-source systems and open-source instruction-tuned models trained on large-scale multilingual corpora. The Vietnamese-focused group comprises models pretrained or fine-tuned on Vietnamese data, including general-purpose Vietnamese chat models and legal domain-adapted LLMs for Vietnamese law. This setup enables a systematic comparison between multilingual modeling and language- and domain-specific adaptation in Vietnamese legal reasoning. Model details are provided in Appendix~\ref{appen:model}.

% \subsection{Main Results}
\begin{table*}[t]
\centering
\scriptsize
\setlength{\tabcolsep}{3.5pt}
\caption{Zero-shot Performance Comparison of Language Models across Different Tasks. \textcolor{red}{\textbf{Red bold}} indicates the best overall performance. \textcolor{blue}{\textbf{Blue bold}} indicates the best performance among open-source models.}
\label{tab:model_comparison}
\begin{tabular}{ll*{10}{c}}
\toprule
& & \multicolumn{5}{c}{\textbf{Recognition \& Recall}} & \multicolumn{5}{c}{\textbf{Understanding \& Structuring}} \\
\cmidrule(lr){3-7} \cmidrule(lr){8-12}
\textbf{Model Type} & \textbf{Model} & \shortstack{1.1\\Acc} & \shortstack{1.2\\Acc} & \shortstack{1.3\\Acc} & \shortstack{1.4\\Acc} & \shortstack{1.5\\Acc} & \shortstack{2.1\\Acc} & \shortstack{2.2\\Acc} & \shortstack{2.3\\R-L} & \shortstack{2.4\\Acc} & \shortstack{2.5\\m-F1} \\
\midrule
\multirow{18}{*}{\textit{\shortstack{General\\Multilingual\\LLMs}}} 
& gpt-4o & 70.03 & 81.14 & 73.67 & 82.50 & 24.25 & 85.37 & 67.33 & 0.470 & 80.67 & \textcolor{red}{\textbf{63.28}} \\
& gpt-4o-mini & 65.24 & 82.71 & 61.33 & 68.40 & 22.25 & \textcolor{red}{\textbf{87.35}} & 51.33 & 0.529 & 73.33 & 61.68 \\
& claude-sonnet-4.5 & 69.78 & 82.47 & \textcolor{red}{\textbf{83.00}} & 84.19 & 27.25 & 79.21 & \textcolor{red}{\textbf{75.33}} & \textcolor{red}{\textbf{0.808}} & \textcolor{red}{\textbf{87.81}} & 62.04 \\
& gemini-2.5-flash & \textcolor{red}{\textbf{71.96}} & 81.40 & 61.33 & 81.40 & 24.25 & 80.63 & 64.00 & 0.656 & 82.30 & 49.92 \\
\cmidrule(lr){2-12}
& gpt-oss-20b & 22.72 & 73.21 & 47.33 & 39.00 & 18.63 & 45.06 & 28.33 & 0.249 & 66.50 & 56.43 \\
& Qwen2.5-72B-Instruct & 68.05 & 80.43 & 77.00 & 79.10 & 21.88 & \textcolor{blue}{\textbf{80.78}} & 65.67 & \textcolor{blue}{\textbf{0.808}} & 83.83 & 23.58 \\
& Qwen2.5-32B-Instruct & 71.25 & \textcolor{red}{\textbf{85.50}} & 71.33 & 79.70 & 22.25 & 79.21 & \textcolor{blue}{\textbf{67.00}} & 0.759 & 80.00 & 56.99 \\
& Qwen2.5-14B-Instruct & 68.58 & 82.28 & 65.67 & 74.60 & 20.50 & 85.38 & 55.53 & 0.733 & 78.17 & 58.96 \\
& Qwen2.5-7B-Instruct & 50.40 & 79.94 & 54.00 & 62.40 & 21.75 & 85.37 & 56.67 & 0.657 & 82.80 & \textcolor{blue}{\textbf{63.28}} \\
& Qwen2.5-3B-Instruct & 52.80 & 70.57 & 50.67 & 57.95 & 25.38 & 72.73 & 48.00 & 0.606 & 67.95 & 51.04 \\
& Llama-3.1-70B-Instruct & 55.68 & 80.67 & 74.33 & 77.79 & 24.38 & 75.68 & 58.67 & 0.516 & 81.80 & 58.12 \\
& Llama-3.1-8B-Instruct & 55.88 & 80.67 & 56.00 & 62.40 & 25.75 & 74.11 & 50.00 & 0.364 & 72.95 & 51.31 \\
& Llama-2-13b-chat-hf & 22.72 & 21.23 & 21.00 & 42.20 & \textcolor{red}{\textbf{27.75}} & 56.10 & 30.24 & 0.411 & 48.00 & 41.81 \\
& Llama-2-7b-chat-hf & 25.52 & 17.28 & 19.67 & 19.20 & 25.50 & 51.22 & 27.02 & 0.198 & 49.33 & 44.63 \\
& gemma-2-27b-it & 57.89 & 78.62 & 58.67 & 72.31 & 24.13 & 73.83 & 56.33 & 0.719 & 83.97 & 55.39 \\
& gemma-2-9b-it & 52.94 & 77.30 & 65.00 & 65.60 & 23.25 & 79.45 & 48.00 & 0.349 & 79.47 & 48.65 \\
& internlm3-8b-instruct & 57.21 & 46.27 & 51.00 & 55.48 & 24.50 & 68.38 & 42.67 & 0.395 & 67.28 & 52.30 \\
& internlm-chat-20b & 16.43 & 24.01 & 17.91 & 21.70 & 18.13 & 11.11 & 9.00 & 0.188 & 61.17 & 32.73 \\
\midrule
\multirow{3}{*}{\textit{\shortstack{Domain-adapted\\Vietnamese LLMs}}}
& SeaLLMs-v3-7B-Chat & 62.23 & 68.96 & 57.67 & 58.88 & 22.50 & 76.28 & 49.67 & 0.475 & 62.10 & 54.39 \\
& SeaLLMs-v3-1.5B-Chat & 47.73 & 49.04 & 55.33 & 39.00 & 25.75 & 42.69 & 27.33 & 0.576 & 48.50 & 47.21 \\
& BloomVN-8B-chat & 46.66 & 65.59 & 63.67 & 65.29 & 26.75 & 70.36 & 45.00 & 0.500 & 49.08 & 57.00 \\
& Qwen3-4b-legal-pretrain  & 62.83 & 75.99 & 66.67 & 70.35 & 24.50 & 79.45 & 45.00 & 0.716 & 73.29 & 58.69 \\
& CMC-Legal-32B & 66.17 & 84.19 & \textcolor{blue}{\textbf{81.67}} & \textcolor{red}{\textbf{87.91}} & 24.13 & \textcolor{blue}{\textbf{80.78}} & 61.33 & 0.694 & \textcolor{blue}{\textbf{87.45}} & 60.58 \\
\bottomrule
\end{tabular}
\end{table*}

\begin{table*}[t]
\centering
\scriptsize
\setlength{\tabcolsep}{3.5pt}
\caption{Zero-shot performance comparison across different tasks. \textcolor{red}{\textbf{Red bold}} denotes the best overall result, while \textcolor{blue}{\textbf{blue bold}} indicates the best-performing open-source model. For Task~3.4, we report F1 scores separately for the Yes and No labels (Y-F1 / N-F1).}
\label{tab:legal_benchmark}
\begin{tabular}{ll*{12}{c}}
\toprule
& & \multicolumn{5}{c}{\textbf{Reasoning \& Infer.}} & \multicolumn{3}{c}{\textbf{Interpret. \& Generation}} & \multicolumn{4}{c}{\textbf{Ethics, Fairness \& Bias}} \\
\cmidrule(lr){3-7} \cmidrule(lr){8-10} \cmidrule(lr){11-14}
\textbf{Model Type} & \textbf{Model} & \shortstack{3.1\\Acc} & \shortstack{3.2\\Acc} & \shortstack{3.3\\Acc} & \shortstack{3.4\\Y-F1/N-F1} & \shortstack{3.5\\Acc} & \shortstack{4.1\\R-L} & \shortstack{4.2\\R-L} & \shortstack{4.3\\R-L} & \shortstack{5.1\\Acc} & \shortstack{5.2\\Acc} & \shortstack{5.3\\Acc} & \shortstack{5.4\\Acc} \\
\midrule
\multirow{16}{*}{\textit{\shortstack{General\\Multilingual\\LLMs}}} 
& gpt-4o & 38.83 & 84.50 & 73.50 & 27.21/42.16 & 67.97 & 0.3257 & 0.4017 & \textcolor{red}{\textbf{0.4975}} & 44.18 & 67.74 & 91.04 & 58.11 \\
& gpt-4o-mini & 35.33 & 82.17 & 74.66 & 11.85/39.59 & 57.38 & \textcolor{red}{\textbf{0.3272}} & 0.4167 & 0.4926 & 41.76 & 67.74 & 86.56 & 51.28 \\
& claude-sonnet-4.5 & \textcolor{red}{\textbf{59.83}} & 88.83 & \textcolor{red}{\textbf{79.10}} & 0/39.61 & \textcolor{red}{\textbf{69.36}} & 0.2842 & 0.3857 & 0.3830 & 45.38 & 72.81 & 91.04 & 69.66 \\
& gemini-2.5-flash & 40.83 & 84.67 & 76.71 & 0/39.61 & 63.51 & 0.2756 & 0.3982 & 0.3778 & 46.18 & 68.66 & 94.04 & 61.54 \\
\cmidrule(lr){2-14}
& gpt-oss-20b & 29.67 & 66.00 & 52.74 & 73.41/35.56 & 37.15 & 0.0262 & 0.1529 & 0.3104 & 21.69 & 58.80 & 33.76 & 28.20 \\
& Qwen2.5-72B-Instruct & 33.50 & 85.50 & 74.32 & 0.00/39.61 & 63.51 & 0.2930 & 0.4076 & \textcolor{blue}{\textbf{0.4825}} & 46.59 & 67.28 & 90.55 & 64.53 \\
& Qwen2.5-32B-Instruct & 32.66 & 82.67 & 74.66 & 0.00/39.61 & 59.78 & 0.3111 & 0.3968 & 0.4652 & 45.78 & \textcolor{red}{\textbf{69.12}} & 92.07 & 66.23 \\
& Qwen2.5-14B-Instruct & 39.67 & 82.17 & 71.92 & 1.59/39.81 & 53.35 & 0.2707 & 0.0168 & 0.4050 & \textcolor{red}{\textbf{57.79}} & 67.59 & 91.54 & 68.38 \\
& Qwen2.5-7B-Instruct & 35.83 & 81.67 & 71.23 & 0.00/39.61 & 56.15 & 0.2531 & 0.3286 & 0.4680 & 36.94 & 60.65 & 89.57 & 68.37 \\
& Qwen2.5-3B-Instruct & 26.67 & 68.67 & 69.86 & 10.61/41.00 & 45.81 & 0.2586 & 0.3605 & 0.4082 & 36.94 & 54.17 & 83.59 & 59.40 \\
& Llama-3.1-70B-Instruct & 35.00 & 86.00 & 75.34 & 0.00/39.61 & 62.11 & 0.3077 & 0.4007 & 0.4005 & 43.77 & 64.05 & \textcolor{red}{\textbf{93.99}} & 55.59 \\
& Llama-3.1-8B-Instruct & 28.00 & 78.67 & 66.44 & 37.66/\textcolor{red}{\textbf{46.07}} & 52.92 & \textcolor{blue}{\textbf{0.3160}} & 0.3776 & 0.3879 & 37.35 & 49.31 & 91.01 & 62.82 \\
& Llama-2-13b-chat-hf & 21.67 & 45.67 & 32.53 & 85.91/0.00 & 30.79 & 0.0149 & 0.0759 & 0.0478 & 21.69 & 28.24 & 27.66 & 27.35 \\
& Llama-2-7b-chat-hf & 21.83 & 29.67 & 23.63 & 33.17/0.00 & 33.62 & 0.0132 & 0.2693 & 0.0318 & 20.08 & 26.85 & 29.17 & 21.37 \\
& gemma-2-27b-it & 31.67 & 81.50 & 73.63 & 0.00/39.61 & 60.45 & 0.2971 & 0.3672 & 0.4654 & 40.16 & 57.60 & 89.07 & 69.66 \\
& gemma-2-9b-it & 28.83 & 80.83 & 72.95 & 62.63/44.78 & 49.44 & 0.3272 & 0.3600 & 0.4678 & 40.16 & 59.26 & 91.05 & 62.82 \\
& internlm3-8b-instruct & 26.67 & 71.00 & 65.07 & 40.96/8.33 & 43.02 & 0.2647 & 0.2694 & 0.2884 & 15.62 & 58.80 & 67.37 & 42.31 \\
& internlm-chat-20b & 19.67 & 37.17 & 23.29 & 64.92/0.00 & 32.59 & 0.0636 & 0.2575 & 0.2917 & 29.31 & 23.96 & 42.79 & 26.07 \\
\midrule
\multirow{3}{*}{\textit{\shortstack{Domain-adapted\\Vietnamese LLMs}}}
& SeaLLMs-v3-7B-Chat & 26.00 & 81.50 & 64.38 & 0.00/39.61 & 49.16 & 0.1700 & 0.3547 & 0.4141 & 39.76 & 56.94 & 92.53 & 63.25 \\
& SeaLLMs-v3-1.5B-Chat & 26.67 & 55.00 & 44.52 & 85.91/0.00 & 34.92 & 0.2180 & 0.2830 & 0.4117 & 25.30 & 31.02 & 68.20 & 57.26 \\
& BloomVN-8B-chat & 32.17 & 82.33 & 70.89 & 0.00/39.61 & 50.00 & 0.2407 & 0.3188 & 0.4099 & 20.10 & 47.22 & 86.61 & 58.55 \\
& Qwen3-4b-legal-pretrain   & \textcolor{blue}{\textbf{43.83}} & 82.00 & 76.37 & 0.00/39.61 & 54.47 & 0.1147 & 0.3737 & 0.4381 & 34.54 & 61.57 & 89.08 & 59.40 \\
& CMC-Legal-32B & 41.67 & \textcolor{red}{\textbf{90.67}} & \textcolor{blue}{\textbf{76.71}} & \textcolor{red}{\textbf{86.41}}/13.33 & \textcolor{blue}{\textbf{62.40}} & 0.2917 & \textcolor{red}{\textbf{0.4213}}& 0.3695 & 32.93 & 60.83 & 92.06 & \textcolor{red}{\textbf{73.50}} \\
\bottomrule
\end{tabular}
\end{table*}

\begin{figure*}[ht]
    \centering
    \includegraphics[width=\textwidth]{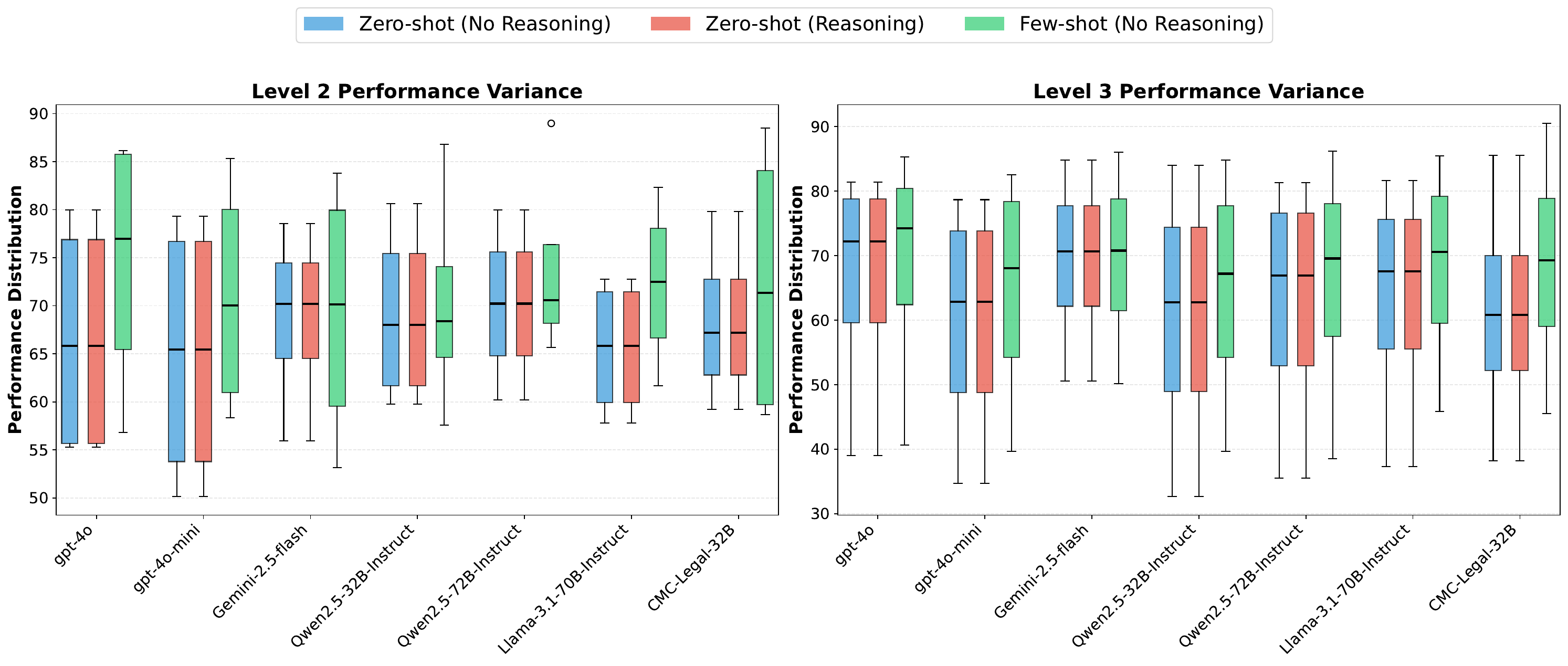}
\caption{\textbf{Ablation of prompting strategies.} Accuracy score distributions across prompting strategies, aggregated over Accuracy-based tasks within each cognitive level (excluding tasks evaluated with other metrics).}
    \label{fig:ablation_distribution}
\end{figure*}

\subsection{Details Performance Analysis}
 
Table~\ref{tab:model_comparison} and Table~\ref{tab:legal_benchmark} present the zero-shot performance of 23 language models across all 22 tasks in VietLegal. We provide additional evaluation results with alternative task formulations in Appendix \ref{appen:additional_result}.   
 
\paragraph{Performance Degradation Across Cognitive Levels}
We observe a dramatic performance decline as tasks increase in cognitive complexity. While top-performing models achieve 80-90\% accuracy on most basic recognition tasks (Level 1), performance drops precipitously on advanced reasoning tasks. Most notably, Task 3.1 (Article/Clause Prediction) proves exceptionally challenging, with the best model achieving only 59.83\% accuracy, a decline of over 28 percentage points compared to the simpler Task 1.4 (Article Recall at 87.91\%). This suggests that while models can recall legal articles when explicitly prompted, applying this knowledge to predict relevant provisions from novel queries remains fundamentally difficult.
 
\paragraph{Systematic Failure in Conflict Detection}
Task 3.4 (Conflict \& Consistency Detection) reveals a systematic failure across nearly all models. Of the 23 models evaluated, 10 achieve a 0.00 Y-F1 score, indicating complete inability to detect legal conflicts. These models exhibit strong bias toward predicting ``no conflict,'' achieving 39.61 N-F1 while failing on positive cases. Only four models demonstrate conflict detection capability on both labels: gpt-oss-20b (73.41/35.56 Y-F1/N-F1), gemma-2-9b-it (62.63/44.78), Llama-3.1-8B-Instruct (37.66/46.07), and GPT-4o (27.21/42.16). These near-constant predictions across most models underscore the difficulty of complex legal reasoning tasks. 
% Additional error analysis for Task~3.4 is presented in Appendix~\ref{appen:error_analysis}.
 
\subsubsection{Domain Adaptation vs. Scale}
 
\paragraph{Impact of Domain Adaptation}
Our results challenge the assumption that larger general-purpose models consistently outperform smaller domain-adapted ones. Vietnamese legal models excel on higher-level cognitive tasks: CMC-Legal-32B achieves state-of-the-art performance on court decision prediction (90.67\%), judicial reasoning generation (0.4213 ROUGE-L), and unfair contract detection (73.50\%), surpassing GPT-4o by 6.17 and 15.39 points on court decision prediction and unfair contract detection. Notably, the 4B-parameter Qwen3-4b-legal-pretrain achieves the best open-source result on article prediction (43.83\%), outperforming models up to 18$\times$ larger, highlighting the effectiveness of domain-specific pretraining over scale.
 
\paragraph{The Diminishing Advantage of Proprietary Models}
While proprietary models (GPT-4o, Claude Sonnet 4.5, Gemini 2.5 Flash) maintain advantages on foundational tasks (Levels 1-2), their superiority diminishes on advanced reasoning and generation tasks (Levels 3-5). GPT-4o achieves best overall performance on only one high-level task: legal opinion generation (0.4975 ROUGE-L). Across six of the remaining ten tasks in Levels 3-5 (excluding conflict detection), domain-adapted Vietnamese models outperform GPT-4o, with particularly large gaps on unfair contract detection (+15.39 points) and court decision prediction (+6.17 points), although Claude Sonnet 4.5 remains the best model on article prediction (59.83\%), multi-article reasoning (79.10\%), and penalty estimation (69.36\%). This pattern suggests that general-purpose training, even at a massive scale, cannot fully substitute for domain-specific legal knowledge when tackling complex juridical reasoning.
 
\subsubsection{Task-Specific Insights}
 
\paragraph{Recognition \& Understanding (Levels 1-2)}
Foundational tasks are generally well handled, with legal topic classification (Task~1.2) reaching 81-86\% accuracy for top models. In contrast, legal schema recall (Task~1.5) remains a universal bottleneck, with all models scoring below 28\%, highlighting persistent difficulties in modeling Vietnamese legal hierarchies, amendments, and temporal relations. Among understanding tasks, Claude Sonnet~4.5 stands out on structured extraction, achieving 0.808 ROUGE-L on legal graph structuring (Task~2.3, tied with Qwen2.5-72B-Instruct) and 87.81\% accuracy on judgment verification (Task~2.4), indicating strong capabilities in complex document analysis.
 
\paragraph{Reasoning \& Inference (Level 3)}
Level~3 tasks reveal sharp capability differences. Court decision prediction (Task~3.2) achieves relatively high accuracy (84-91\%), suggesting strong pattern-based reasoning, whereas article prediction from short queries (Task~3.1) remains challenging (19.67-59.83\%). Penalty estimation (Task~3.5) further exposes architectural gaps: Claude Sonnet~4.5 achieves the highest accuracy (69.36\%), followed by GPT-4o (67.97\%), outperforming all open-source models, indicating that nuanced discretionary judgment remains difficult for current non-proprietary systems.
 
\paragraph{Generation \& Ethics (Levels 4-5)}
Generation tasks show moderate performance overall (best ROUGE-L 0.33-0.50), with legal opinion generation (Task~4.3) more tractable than summarization or IRAC-style reasoning. Ethics and fairness tasks exhibit a clear split: models achieve high ethical consistency (Task~5.3, often $>85\%$) but struggle with bias detection (Task~5.1, 15-58\%). Notably, Qwen2.5-14B achieves the strongest bias detection performance (57.79\%), surpassing larger models, suggesting that sensitivity to legal bias depends more on training characteristics than model scale.
 
\subsubsection{Model Family Analysis}
 
\paragraph{Qwen2.5 Series.} The Qwen2.5 family shows consistent performance across sizes, with the 32B model achieving the best legal topic classification (85.50\%). However, the 72B, 32B, and 7B variants completely fail at conflict detection (0.00 Y-F1), and the 14B and 3B variants nearly fail (1.59 and 10.61 Y-F1), indicating a likely systematic limitation. Despite this, the family demonstrates strong generation ability (0.41--0.48 ROUGE-L on Task~4.3) and competitive results on most classification and reasoning tasks, making it a solid baseline for Vietnamese legal AI. 
 
\textbf{Llama Series.} The Llama-3.1 models exhibit mixed performance. Llama-3.1-70B achieves the highest ethical consistency among open-source models (93.99\%), while the 8B variant is one of the few models showing meaningful conflict detection (37.66/46.07 Y-F1/N-F1). This suggests notable architectural differences. In contrast, Llama-2 models perform poorly across advanced tasks, with accuracy often below 30\% and ROUGE-L mostly under 0.10, highlighting rapid progress in domain-specialized LLMs. 
 
\textbf{Domain-Adapted Vietnamese Models.} The Vietnamese domain-adapted models illustrate clear gains from specialization. BloomVN-8B-chat delivers moderate performance comparable to general multilingual models. CMC-Legal-32B, with legal-specific fine-tuning, dominates reasoning and inference tasks among open-source models, achieving the best overall result on court decision prediction and the best open-source result on multi-article reasoning. This progression from general Vietnamese to legal-specialized models supports a staged approach to developing effective legal AI for low-resource languages.
 
\subsubsection{Key Takeaways}
 
Our experimental results yield four primary conclusions: (1) Current LLMs face fundamental challenges in advanced legal reasoning, with article prediction and conflict detection remaining largely unsolved; (2) Domain-specific pretraining provides greater benefits than parameter scaling for complex legal tasks, as evidenced by small specialized models outperforming general models up to 18 times their size; (3) Proprietary model advantages diminish on high-level legal reasoning tasks, where domain-adapted models outperform GPT-4o on most tasks; and (4) Certain capabilities particularly bias detection and legal schema understanding remain challenging for all models, representing critical areas for future research in legal AI.
 
\section{Contamination Study}
To assess data contamination, we analyzed 1000 stratified instances across all task categories using complementary detection methods, including n-gram web search, targeted Vietnamese legal portal screening, and Common Crawl verification. Potential matches were further filtered to distinguish substantive case content from mandatory statutory text and standard legal templates. Across all methods, only 1.8\% of instances showed potential overlap, all attributable to inherently duplicative statutory provisions or templates by design, with no contamination observed in case-based, reasoning, or generation tasks. These results indicate minimal contamination risk; full methodological details are provided in Appendix~\ref{appen:contamination}.

\section{Error Analysis}
\label{sec:error-analysis}

To move beyond leaderboard rankings and identify fundamental bottlenecks of LLMs in Vietnamese legal reasoning, we conduct a systematic error analysis across six representative tasks covering all five cognitive levels, with particular emphasis on tasks where most models achieve low performance.

\paragraph{Level 1 (Task 1.5).}
Three consistent failure modes emerge in Legal Schema Recall. Models exhibit hierarchical document-type bias, defaulting to higher-level instruments (e.g., Law instead of the correct Circular) due to rigid assumptions about the Vietnamese legal hierarchy. They also demonstrate temporal-direction confusion, frequently misinterpreting the direction of amendment relationships when the amending instrument carries a contextually unexpected issuance date. Finally, consolidation confusion is pervasive: models conflate Consolidation with replacement or annulment, failing to distinguish the procedural role of a Consolidated Document.

\paragraph{Level 3 (Tasks 3.1 \& 3.4).}
In Article/Clause Prediction, two compounding errors arise. Recency blindness causes models to select outdated legislation over correct 2024-2025 instruments, questions involving the Law on Road Traffic Order and Safety 2024, the Law on Organization of People's Courts 2024, and Decree No.~37/2025/ND-CP were answered incorrectly by all evaluated models. Sub-article granularity failure further compounds this, as models often select the wrong Article or Clause among closely numbered distractors even when the correct statute is identified. In Conflict \& Consistency Detection, two mirror-image biases emerge: general-purpose models exhibit a no-conflict bias, returning false negatives for genuine contradictions between provincial decisions and national statutes, while CMC-Legal-32B shows over-detection bias, predicting conflict in nearly all instances, including clearly consistent pairs. Both failure patterns confirm that cross-document normative reasoning over hierarchical norms and temporal validity remains unresolved across all 23 evaluated models.

\paragraph{Level 4 (Tasks 4.1, 4.2 \& 4.3).}
Across generation tasks, two cross-cutting weaknesses emerge: format non-compliance and insufficient statutory grounding. In Legal Document Summarization, models replicate the source document's organizational structure rather than producing abstractive summaries, and outputs are frequently truncated mid-sentence. In Judicial Reasoning Generation, the Rule component of the IRAC framework is most error-prone: models fabricate non-existent legislation and systematically misidentify jurisdiction-specific thresholds, such as defaulting to the UN Convention's under-18 definition rather than the correct under-16 threshold in the Law on Children 2016. In Objective Legal Opinion Generation, models introduce external references absent from the provided text and frequently violate explicit formatting constraints despite clear instructions.

\paragraph{Level 5 (Task 5.1).}
A systematic single-answer truncation pattern dominates this task. When the correct answer requires multiple selections, models output only the most salient option, missing subtler discrimination forms such as positive stereotyping or implicit favoritism. GPT-4o and GPT-4o-mini exhibit this pattern across most multi-answer questions, while CMC-Legal-32B produces frequent empty predictions, reflecting overly cautious generation misaligned with the task's ethical reasoning format.

\section{Ablation Study}

\paragraph{Experimental Settings.}
We systematically evaluate large language models on the Vietnamese Legal Benchmark under multiple controlled settings to isolate the effects of prompting strategies and external knowledge access. Specifically, we consider: (i) \textbf{Zero-shot with explicit reasoning}, where models are instructed to generate intermediate chain-of-thought explanations; (ii) \textbf{Few-shot without reasoning}, where models are provided with a small number of task demonstrations but no explicit reasoning requirement; and (iii) \textbf{Agentic Retrieval-Augmented Generation}, where models iteratively retrieve relevant statutory documents from Legal Corpus Database using a search tool before producing answers. For generation tasks, we additionally conduct double-blind \textbf{human evaluation} with legal experts, enabling direct comparison of prompting and retrieval strategies under identical task and metric conditions. For completeness, detailed task-level results, additional ablation settings, and extended analyses are reported in Appendix~\ref{appen:more_result}.

\paragraph{Prompting strategies.}
Figure~\ref{fig:ablation_distribution} summarizes the impact of prompting strategies by visualizing performance distributions across tasks within each evaluation level. Each box plot aggregates task-level scores for a given model, enabling comparison of both central tendency and performance stability. Across models, few-shot prompting without explicit reasoning consistently yields higher and more stable performance than zero-shot prompting with reasoning. This effect is especially pronounced for Understanding \& Structuring tasks (Level~2), where explicit reasoning often degrades performance and increases variance, reflecting the reliance of these tasks on holistic pattern recognition rather than step-by-step decomposition. For Reasoning \& Inference tasks (Level~3), explicit reasoning shows mixed and task-dependent effects, benefiting some procedural tasks such as article identification but remaining unstable for nuanced judgments like conflict detection, whereas few-shot demonstrations consistently reduce variance and improve median performance by implicitly inducing task-appropriate reasoning behaviors.

\section{Related Works}

\paragraph{Legal LLM Benchmarks.}
Early legal NLP benchmarks primarily targeted isolated tasks such as judgment prediction or statute classification, exemplified by CaseHOLD \cite{zheng2021}. More recent efforts have shifted toward multi-task evaluations of general legal intelligence, most notably LexGLUE \cite{chalkidis2022} and LegalBench \cite{guha2023}, which emphasize legal reasoning beyond surface-level language understanding. Parallel developments include benchmarks for legally specialized LLMs \cite{cui2023,yue2023} and civil law-oriented resources, particularly for Chinese \cite{fei2024lawbench,dai2025laiw}. European civil law benchmarks \cite{merryman2018civil}, such as French statutory retrieval \cite{louis2022french} and German civil law QA \cite{buttner2024german}, further highlight structural differences in codified legal systems. Despite this progress, low-resource languages and many civil law jurisdictions in the Global South remain underrepresented, leaving Vietnamese law largely unexplored in existing benchmark landscapes.

\paragraph{Vietnamese Legal NLP.}
Vietnamese legal NLP research has been driven largely by community-led shared tasks, particularly through the VLSP workshops \cite{luu2025vlsp,tran2024vlsp}, which have produced datasets for legal retrieval, entailment, and question answering. Pre-trained models such as PhoBERT \cite{nguyen2020} and ViT5 \cite{phan2022} have enabled strong performance on these foundational tasks. However, existing resources are fragmented and predominantly retrieval- or extraction-focused, offering limited evaluation of generative reasoning, multi-step inference, or legislative amendment tracking. Recent work on Vietnamese legal RAG systems \cite{nguyen2024} further underscores the lack of a unified, standardized benchmark capable of evaluating realistic legal assistant workflows.

\paragraph{Cognitive Evaluation and Metrics.}
Recent benchmarking efforts increasingly draw on cognitive frameworks to distinguish memorization from higher-order reasoning. Bloom's Taxonomy has been adopted to structure task difficulty, while Chain-of-Thought prompting \cite{wei2022} has emphasized the importance of evaluating intermediate reasoning. In legal NLP, evaluation remains challenging, as standard generation metrics often correlate poorly with factual correctness \cite{liu2023}. We therefore adopt a hybrid evaluation strategy, combining extraction-based metrics for lower cognitive levels with generation metrics for higher-level tasks. Crucially, legal reasoning is grounded in legal syllogism and subsumption theory \cite{adler1989theory}, where correctly applying statutory norms to factual scenarios is central.

\section{Conclusion}
Based on our analyses and experiments, \textbf{VietLegal} advances the evaluation of large language models in the Vietnamese legal domain by introducing a \textit{civil-law-oriented, cognition-based benchmark} tailored to Vietnam’s codified legal system. Through hierarchical statutory structures, scenario-based tasks, and a Bloom’s taxonomy-driven framework, it enables systematic assessment from factual recall to complex legal reasoning. The release of 10,450 expert-verified samples improves reproducibility and supports extension to other civil law jurisdictions. Experiments show that while current LLMs perform well on lower-level tasks, they struggle with advanced reasoning and cross-statutory interpretation, underscoring the limits of general-purpose models. Overall, VietLegal provides a robust evaluation standard and a foundation for future research on legally reliable LLMs in Vietnamese and other civil law systems.

\newpage
\bibliographystyle{plainnat}
\bibliography{citation}
%%
%% If your work has an appendix, this is the place to put it.
\newpage
\section*{NeurIPS Paper Checklist}

\begin{enumerate}

\item {\bf Claims}
    \item[] Question: Do the main claims made in the abstract and introduction accurately reflect the paper's contributions and scope?
    \item[] Answer: \answerYes{} % Replace by \answerYes{}, \answerNo{}, or \answerNA{}.
    \item[] Justification: The main claims made in the abstract and introduction accurately reflect the main paper's contributions and scope.

\item {\bf Limitations}
    \item[] Question: Does the paper discuss the limitations of the work performed by the authors?
    \item[] Answer: \answerYes{} % Replace by \answerYes{}, \answerNo{}, or \answerNA{}.
    \item[] Justification:  We include a dedicated limitation discussion in Appendix \ref{appen:limitations}

\item {\bf Theory assumptions and proofs}
    \item[] Question: For each theoretical result, does the paper provide the full set of assumptions and a complete (and correct) proof?
    \item[] Answer: \answerNA{} % Replace by \answerYes{}, \answerNo{}, or \answerNA{}.
    \item[] Justification: There are no theoretical results in this paper.

    \item {\bf Experimental result reproducibility}
    \item[] Question: Does the paper fully disclose all the information needed to reproduce the main experimental results of the paper to the extent that it affects the main claims and/or conclusions of the paper (regardless of whether the code and data are provided or not)?
    \item[] Answer: \answerYes{} % Replace by \answerYes{}, \answerNo{}, or \answerNA{}.
    \item[] Justification: The paper fully discloses all the information needed to reproduce the main experimental results of the paper in Section \ref{sec:full_experiment}, and we provide the code to assist the reproducibility
of our experimental results in \url{https://github.com/CMC-OPENAI/VLegal-Bench}.

\item {\bf Open access to data and code}
    \item[] Question: Does the paper provide open access to the data and code, with sufficient instructions to faithfully reproduce the main experimental results, as described in supplemental material?
    \item[] Answer: \answerYes{} % Replace by \answerYes{}, \answerNo{}, or \answerNA{}.
    \item[] Justification: This paper provides open access to the data and code, and includes instructions for running the code in the link \url{https://github.com/CMC-OPENAI/VLegal-Bench} and the data link in \url{ https://huggingface.co/datasets/CMC-OPENAI/VLegal-Bench}.

\item {\bf Experimental setting/details}
    \item[] Question: Does the paper specify all the training and test details (e.g., data splits, hyperparameters, how they were chosen, type of optimizer) necessary to understand the results?
    \item[] Answer: \answerYes{} % Replace by \answerYes{}, \answerNo{}, or \answerNA{}.
    \item[] Justification: Section \ref{sec:full_experiment} contains all experimental details.

\item {\bf Experiment statistical significance}
    \item[] Question: Does the paper report error bars suitably and correctly defined or other appropriate information about the statistical significance of the experiments?
    \item[] Answer: \answerYes{} % Replace by \answerYes{}, \answerNo{}, or \answerNA{}.
    \item[] Justification: We provide statistical significance analyses of prompt sensitivity in Section \ref{sec:full_experiment}.

\item {\bf Experiments compute resources}
    \item[] Question: For each experiment, does the paper provide sufficient information on the computer resources (type of compute workers, memory, time of execution) needed to reproduce the experiments?
    \item[] Answer: \answerYes{} % Replace by \answerYes{}, \answerNo{}, or \answerNA{}.
    \item[] Justification: We include all experiments computed in the Appendix \ref{appen:compute_resources}, covering all resources used for generating model responses.
    
\item {\bf Code of ethics}
    \item[] Question: Does the research conducted in the paper conform, in every respect, with the NeurIPS Code of Ethics \url{https://neurips.cc/public/EthicsGuidelines}?
    \item[] Answer: \answerYes{} % Replace by \answerYes{}, \answerNo{}, or \answerNA{}.
    \item[] Justification: We confirm the research conducted in the paper conforms, in every respect, with the NeurIPS Code of Ethics.

\item {\bf Broader impacts}
    \item[] Question: Does the paper discuss both potential positive societal impacts and negative societal impacts of the work performed?
    \item[] Answer: \answerYes{} % Replace by \answerYes{}, \answerNo{}, or \answerNA{}.
    \item[] Justification: We discuss broader impact in Appendix \ref{appen:broader_impact}.
    
\item {\bf Safeguards}
    \item[] Question: Does the paper describe safeguards that have been put in place for responsible release of data or models that have a high risk for misuse (e.g., pre-trained language models, image generators, or scraped datasets)?
    \item[] Answer: \answerYes{} % Replace by \answerYes{}, \answerNo{}, or \answerNA{}.
    \item[] Justification: We discuss safeguards that have been put in place for the responsible release of data in Section \ref{sec:data_colection}.

\item {\bf Licenses for existing assets}
    \item[] Question: Are the creators or original owners of assets (e.g., code, data, models), used in the paper, properly credited and are the license and terms of use explicitly mentioned and properly respected?
    \item[] Answer: \answerYes{} % Replace by \answerYes{}, \answerNo{}, or \answerNA{}.
    \item[] Justification: All evaluated models are properly cited with their original papers, and we use the official Hugging Face or API releases of each model under their respective licenses. Full asset details, versions, and URLs are provided in Appendix~\ref{appen:model}.

\item {\bf New assets}
    \item[] Question: Are new assets introduced in the paper well documented and is the documentation provided alongside the assets?
    \item[] Answer: \answerYes{} % Replace by \answerYes{}, \answerNo{}, or \answerNA{}.
    \item[] Justification:  We document all assets.

\item {\bf Crowdsourcing and research with human subjects}
    \item[] Question: For crowdsourcing experiments and research with human subjects, does the paper include the full text of instructions given to participants and screenshots, if applicable, as well as details about compensation (if any)? 
    \item[] Answer: \answerYes{} % Replace by \answerYes{}, \answerNo{}, or \answerNA{}.
    \item[] Justification: We include details for human annotations in Appendix \ref{appen:labelling}.

\item {\bf Institutional review board (IRB) approvals or equivalent for research with human subjects}
    \item[] Question: Does the paper describe potential risks incurred by study participants, whether such risks were disclosed to the subjects, and whether Institutional Review Board (IRB) approvals (or an equivalent approval/review based on the requirements of your country or institution) were obtained?
    \item[] Answer: \answerNA{} % Replace by \answerYes{}, \answerNo{}, or \answerNA{}.
    \item[] Justification: This work does not involve research with human subjects or crowdsourcing of personal data. The annotation process was conducted by professional legal experts (a senior lawyer and trained junior legal annotators) employed for the task, and all source materials are publicly available statutory documents from official Vietnamese government portals. 

\item {\bf Declaration of LLM usage}
    \item[] Question: Does the paper describe the usage of LLMs if it is an important, original, or non-standard component of the core methods in this research? Note that if the LLM is used only for writing, editing, or formatting purposes and does \emph{not} impact the core methodology, scientific rigor, or originality of the research, declaration is not required.
    %this research? 
    \item[] Answer: \answerYes{} % Replace by \answerYes{}, \answerNo{}, or \answerNA{}.
    \item[] Justification: LLMs are the central object of study in this paper, as VietLegal is explicitly designed to evaluate the legal reasoning capabilities of large language models. The 23 models assessed, their versions, training paradigms, and access methods are documented in Section~4.2 and Appendix~\ref{appen:model}, and all evaluation prompts and inference configurations are reported in Appendix~\ref{appen:task instruction} and Appendix~\ref{appen:compute_resources}. LLMs were not used to generate, annotate, or alter any benchmark instance; all 10{,}450 samples were constructed and validated by human legal experts. Any use of LLMs for writing or editing assistance was limited to surface-level language polishing and did not affect the methodology, experimental design, or scientific claims of this work.

\end{enumerate}
\newpage
\clearpage
\appendix
\section*{Appendices}
\addcontentsline{toc}{section}{Appendices}  % optional: adds to main TOC
\startcontents[appendices]
\printcontents[appendices]{}{1}{\setcounter{tocdepth}{2}}

\section{Limitations}\label{appen:limitations}
While VietLegal represents an initial step toward a comprehensive evaluation framework for Vietnamese civil law, several limitations should be acknowledged.

\paragraph{Temporal Validity and Legislative Updates}
Law is inherently dynamic, and Vietnamese legislation is frequently amended, replaced, or repealed. Although VietLegal accounts for legislative relationships through a Knowledge Graph and tracks historical versions at the time of construction, the benchmark itself is static. As a result, newly promulgated or revised statutes (e.g., major legislative changes such as the 2024 Land Law) may render some benchmark items outdated over time. This temporal mismatch limits the long-term validity of fixed benchmark instances and highlights the challenge of maintaining alignment with the current legal corpus.

\paragraph{Linguistic and Structural Nuances}
Vietnamese legal texts employ highly formal language and deeply nested hierarchical structures, which VietLegal is designed to evaluate. However, the benchmark primarily focuses on codified written law. In practice, legal interpretation often relies on administrative circulars, guidance documents, and implicit norms that clarify or, in some cases, complicate the application of higher-level statutes. These sources are not always consistently codified or linguistically explicit, and as a result, the dataset may not fully capture such interpretive gray areas or conflicts present in real-world legal reasoning.

\section{Broader Impact}\label{appen:broader_impact}

VietLegal provides an expert-validated benchmark for evaluating legal reasoning in large language models, supporting the development of more reliable and accessible legal AI systems. By covering a spectrum of competencies from statutory recall to ethical reasoning, it enables more realistic assessment of models in legal assistant scenarios and can help improve access to legal information, reduce barriers to legal literacy, and assist practitioners in routine tasks. While grounded in Vietnamese law, the benchmark design is transferable to other jurisdictions, particularly civil-law systems.

At the same time, risks remain. Strong benchmark performance does not imply readiness for real-world legal deployment, and misuse could lead to incorrect or harmful advice. Models may also inherit biases from training data, raising fairness concerns in sensitive contexts. To mitigate these risks, we position VietLegal strictly as a research and evaluation tool, incorporate dedicated ethics-focused tasks, and emphasize the necessity of human expert oversight in any high-stakes application.

\section{Construct Validity and Robustness Against Pattern Matching}\label{appen:construct_validity}

A common concern in multiple-choice question (MCQ) benchmarks is that models may exploit superficial cues rather than demonstrate genuine reasoning ability. In particular, surface-form features such as answer length, lexical overlap, or statistical regularities can enable pattern-matching strategies that bypass true understanding. We explicitly design VietLegal to mitigate such risks and empirically verify that these shortcuts are ineffective.

\paragraph{Expert-Curated Distractor Design.}
All distractor options in VietLegal are manually authored by licensed legal professionals, rather than automatically generated. This allows precise control over their structure and difficulty. Specifically, distractors are constructed to be (i) \textit{legally plausible}, reflecting realistic misconceptions or confusions between related statutory provisions; (ii) \textit{surface-similar} to correct answers, sharing terminology, phrasing, and article references to prevent keyword-based elimination; and (iii) \textit{cognitively targeted}, particularly at higher difficulty levels, where distractors encode alternative but incorrect legal interpretations that require multi-step reasoning to refute. 

This design ensures that models relying on shallow heuristics are systematically misled, as incorrect options are intentionally crafted to resemble correct ones both lexically and structurally.

\paragraph{Baseline Evaluation of Surface Heuristics.}
To further assess whether the MCQ format can be exploited by non-reasoning strategies, we evaluate three representative baselines: (1) random guessing, (2) a longest-answer heuristic, and (3) a semantic similarity method using PhoBERT, a Vietnamese pretrained encoder.

\begin{table}[h]
\centering
\small
\caption{Performance (\%) of non-reasoning baselines across difficulty levels. Level 4 is excluded as it consists of open-ended generation tasks.}
\label{tab:heuristics}
\begin{tabular}{lcccc}
\toprule
\textbf{Strategy} & \textbf{Level 1} & \textbf{Level 2} & \textbf{Level 3} & \textbf{Level 5} \\
\midrule
Random Guessing & 25.00 & 25.17 & 26.71 & 25.00 \\
Longest Answer & 20.50 & 25.00 & 23.29 & 24.60 \\
PhoBERT Similarity & 34.00 & 28.00 & 29.52 & 33.50 \\
\bottomrule
\end{tabular}
\end{table}

The results in Table~\ref{tab:heuristics} show that all three strategies perform at or near random chance. Notably, the longest-answer heuristic performs below chance at multiple levels, indicating that distractor construction actively penalizes such shortcuts. Meanwhile, semantic similarity with PhoBERT \cite{nguyen2020} achieves only marginal improvements over random guessing, despite being trained on Vietnamese text and exposed to legal language.

\paragraph{Implications for Construct Validity.}
The consistently weak performance of these baselines demonstrates that VietLegal cannot be effectively solved using simple surface-level strategies. In particular, the gap between PhoBERT-based similarity and the significantly higher performance of large language models, details in Section \ref{sec:full_experiment} suggests that successful performance requires qualitatively different capabilities beyond lexical matching. Furthermore, this gap becomes more pronounced at higher difficulty levels, aligning with increased reasoning demands.

Overall, these findings provide empirical support for the construct validity of VietLegal: model performance on the benchmark reflects genuine differences in legal reasoning ability rather than artifacts of the evaluation format.

\begin{figure*}
\includegraphics[width=\textwidth]{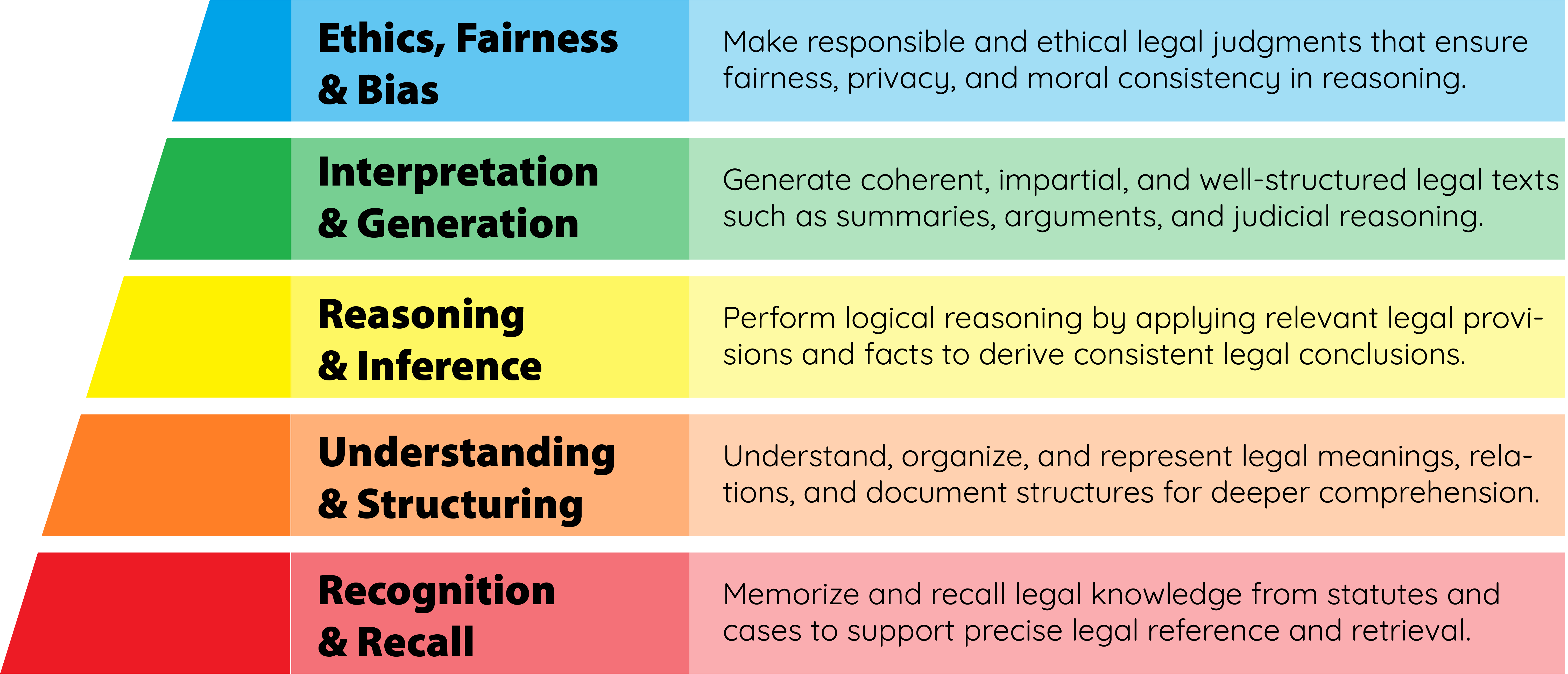}
\caption{The five-level cognitive framework of VietLegal}
\end{figure*}

\section{Computational Resources}\label{appen:compute_resources}
All experiments on open-source and domain-adapted Vietnamese LLMs were conducted on a single on-premise server equipped with 8$\times$
NVIDIA L40 GPUs, 512 GB of system RAM running Ubuntu 22.04 with CUDA 12.1. Local inference was served via \texttt{vLLM} with tensor parallelism scaled to model size. Decoding temperature was fixed at 0 for reproducibility. A single full-benchmark pass (10{,}450 instances, one prompting configuration) required approximately 1 hours for sub-4B models, 2-3 hours for 7-14B models, 3-4 hours for 20-32B models, and 5-6 hours for 70B+ models. Across all 19 locally served models and 4 prompting configurations, the experiments reported in Tables~\ref{tab:model_comparison} and~\ref{tab:legal_benchmark}. Proprietary models (\texttt{gpt-4o}, \texttt{gpt-4o-mini}, \texttt{claude-sonnet-4.5}, and \texttt{gemini-2.5-flash}) were accessed via official APIs. 

\section{Additional Evaluation Formulation}\label{appen:additional_result}
We conducted additional evaluations beyond MCQ accuracy for selected tasks. All distractors in the original MCQ formulation were manually designed by licensed legal experts to be legally plausible and surface-similar to correct answers. The results below provide complementary insights into model capabilities at different abstraction levels.

%% ────────────────────────────────────────────
\subsection{Legal Entity Recognition}
%% ────────────────────────────────────────────

We evaluated entity extraction using standard NER metrics, detailed in Table \ref{tab:task1.1-ner}. The near-zero Exact Match scores indicate that models often identify individual entities but fail to recover the complete entity set required for legal information extraction. 

\begin{table}[h]
\centering
\small
\caption{NER evaluation results for Task 1.1 - Legal Entity Recognition.}
\label{tab:task1.1-ner}
\begin{tabular}{lccc}
\toprule
\textbf{Model} & \textbf{Recall} & \textbf{Precision} & \textbf{Exact Match} \\
\midrule
Claude-Sonnet-4.5         & 0.5748 & 0.5417 & 0.0546 \\
GPT-4o                    & 0.5372 & 0.5904 & 0.0853 \\
GPT-4o-mini               & 0.4914 & 0.6402 & 0.0986 \\
Qwen2.5-32B-Instruct      & 0.4687 & 0.4754 & 0.0375 \\
CMC-Legal-32B     & 0.5810 & 0.5872 & 0.1184 \\
\bottomrule
\end{tabular}
\end{table}

%% ────────────────────────────────────────────
\subsection{Legal Topic Classification}
%% ────────────────────────────────────────────

We reformulated Task 1.2 as a standard classification problem and computed Macro-F1, detailed in Table \ref{tab:task1.2-f1}. Macro-F1 highlights stronger differences between models and reveals that several models largely default to majority-class predictions.

\begin{table}[h]
\centering
\small
\caption{Macro-F1 evaluation results for Task 1.2 - Legal Topic Classification.}
\label{tab:task1.2-f1}
\begin{tabular}{lc}
\toprule
\textbf{Model} & \textbf{Macro-F1} \\
\midrule
GPT-4o                    & 0.8115 \\
GPT-4o-mini               & 0.8284 \\
CMC-Legal-32B     & 0.8560 \\
Qwen2.5-32B-Instruct      & 0.8560 \\
Qwen2.5-14B-Instruct      & 0.8247 \\
Qwen2.5-7B-Instruct       & 0.7977 \\
Qwen2.5-3B-Instruct       & 0.6977 \\
GPT-oss-20B               & 0.7480 \\
Gemma-2-27B-Instruct      & 0.7855 \\
Gemma-2-9B-Instruct       & 0.7725 \\
Llama-3.1-70B-Instruct    & 0.8057 \\
Llama-2-13B-chat          & 0.1479 \\
Llama-2-7B-chat           & 0.0573 \\
SeaLLMs-v3-7B-Chat        & 0.6872 \\
SeaLLMs-v3-1.5B-Chat      & 0.4791 \\
BloomVN-8B-chat           & 0.6517 \\
InternLM3-8B-Instruct     & 0.3995 \\
InternLM-chat-20B         & 0.1601 \\
\bottomrule
\end{tabular}
\end{table}

%% ────────────────────────────────────────────
\subsection{Legal Knowledge Graph Structuring}
%% ────────────────────────────────────────────

We evaluated node and edge prediction quality separately, detailed in Table \ref{tab:task2.3-kg}. These metrics reveal that models generally perform better at predicting relational edges than correctly identifying the corresponding legal nodes (e.g., specific articles or clauses), suggesting that models capture relational legal patterns but struggle with precise legal reference extraction.

\begin{table}[h]
\centering
\small
\caption{Node-F1 and Edge-F1 evaluation results for Task 2.3 - Legal Knowledge Graph Structuring.}
\label{tab:task2.3-kg}
\begin{tabular}{lcc}
\toprule
\textbf{Model} & \textbf{Node-F1} & \textbf{Edge-F1} \\
\midrule
GPT-4o                    & 0.2037 & 0.7989 \\
GPT-4o-mini               & 0.1822 & 0.5390 \\
CMC-Legal-32B     & 0.0412 & 0.8352 \\
Qwen2.5-72B-Instruct      & 0.0296 & 0.7325 \\
Qwen2.5-32B-Instruct      & 0.4250 & 0.8216 \\
Qwen2.5-14B-Instruct      & 0.1511 & 0.7869 \\
Qwen2.5-7B-Instruct       & 0.0419 & 0.3363 \\
Qwen2.5-3B-Instruct       & 0.1145 & 0.3622 \\
GPT-oss-20B               & 0.0074 & 0.0761 \\
Gemma-2-27B-Instruct      & 0.3508 & 0.4698 \\
Gemma-2-9B-Instruct       & 0.0370 & 0.2613 \\
Llama-3.1-70B-Instruct    & 0.0457 & 0.4361 \\
Llama-3.1-8B-Instruct     & 0.0495 & 0.2772 \\
Llama-2-13B-chat          & 0.0295 & 0.0328 \\
Llama-2-7B-chat           & 0.0234 & 0.0128 \\
SeaLLMs-v3-7B-Chat        & 0.0573 & 0.0675 \\
SeaLLMs-v3-1.5B-Chat      & 0.0656 & 0.0725 \\
BloomVN-8B-chat           & 0.0577 & 0.2416 \\
InternLM3-8B-Instruct     & 0.0000 & 0.0000 \\
InternLM-chat-20B         & 0.0194 & 0.0110 \\
\bottomrule
\end{tabular}
\end{table}
\subsection{BERTScore Evaluation for Generation Tasks}
\label{appen:bertscore}

In addition to ROUGE-L, we report BERTScore F1 using \texttt{bert-base-multilingual-cased} to better capture semantic similarity beyond surface-level lexical overlap. This metric is particularly suitable for legal text, where correct answers may involve substantial paraphrasing while preserving meaning.

\begin{table}[h]
\centering
\small
\caption{BERTScore F1 for open-ended generation tasks.}
\label{tab:bertscore}
\begin{tabular}{lccc}
\toprule
\textbf{Model} & \textbf{Task 4.1} & \textbf{Task 4.2} & \textbf{Task 4.3} \\
\midrule
Gemini-2.5-Flash & 0.7053 & 0.7429 & 0.7391 \\
Claude-Sonnet-4.5 & 0.7253 & 0.7318 & 0.7397 \\
Qwen2.5-72B-Instruct & 0.7210 & 0.7454 & 0.7643 \\
Qwen2.5-32B-Instruct & 0.7102 & 0.7406 & 0.7560 \\
Qwen2.5-14B-Instruct & 0.7048 & 0.5033 & 0.7469 \\
Llama-3.1-8B-Instruct & 0.7335 & 0.7191 & 0.7301 \\
Gemma-2-9B-Instruct & 0.7036 & 0.7351 & 0.7593 \\
Gemma-2-27B-Instruct & 0.7045 & 0.7396 & 0.7580 \\
\bottomrule
\end{tabular}
\end{table}

Overall, model rankings under BERTScore are broadly consistent with those observed using ROUGE-L, supporting the use of ROUGE-L as a reliable metric for relative comparison in our setting. Among the three tasks, Task 4.2 (Judicial Reasoning Generation) exhibits the greatest performance variance, aligning with human evaluation results that identify it as the most challenging generation task.

A notable outlier is Qwen2.5-14B-Instruct, which shows a substantial drop in Task 4.2 compared to its performance on the other tasks. This suggests a specific weakness in structured legal reasoning, particularly in generating coherent IRAC-style arguments. Importantly, this limitation is reflected consistently across both BERTScore and human evaluation, providing converging evidence for the robustness of our evaluation framework.
\section{Details of Annotation System}\label{appen:annotate}
\begin{figure*}[t]
    \includegraphics[width=\textwidth]{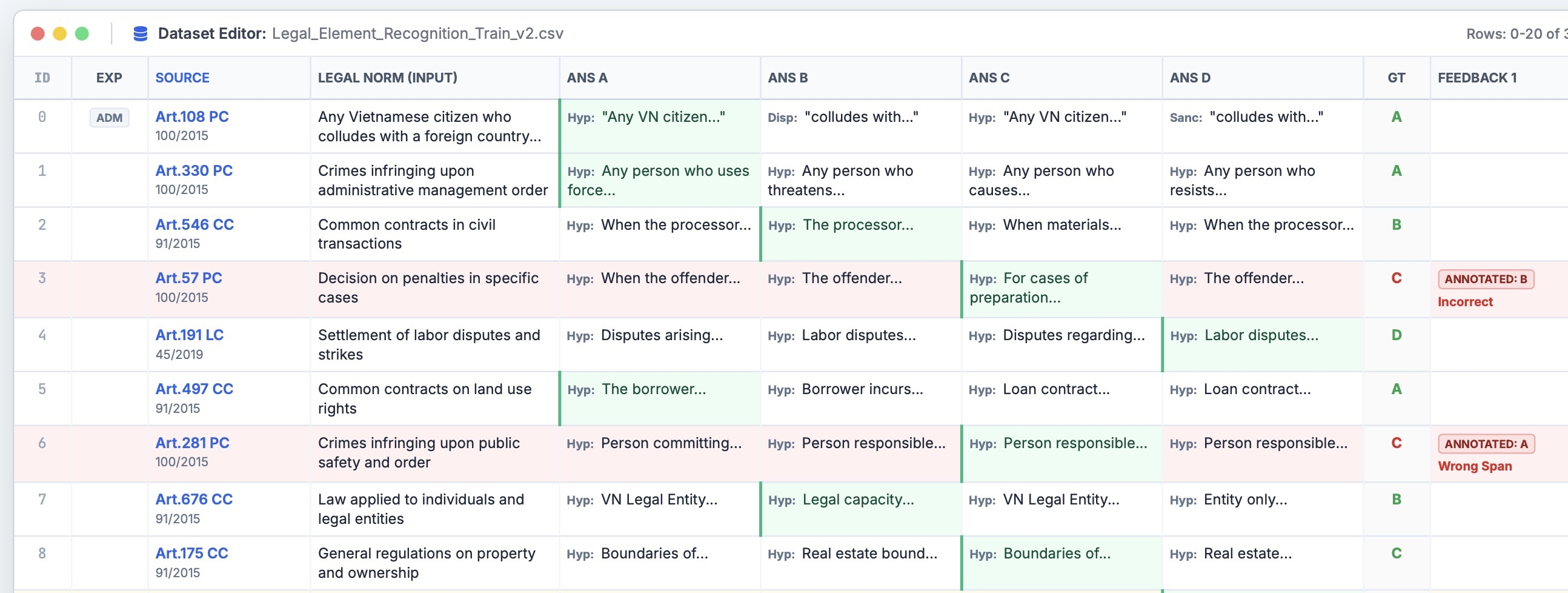}
    \caption{Annotation tool interface: a custom-built tool that supports junior annotators by attaching senior-selected Articles, Clauses, and Points. Junior experts create samples and perform cross-verification in the feedback column.}
    \label{fig:annotation_tool}
\end{figure*}

\begin{figure*}[t]
    \includegraphics[width=\textwidth]{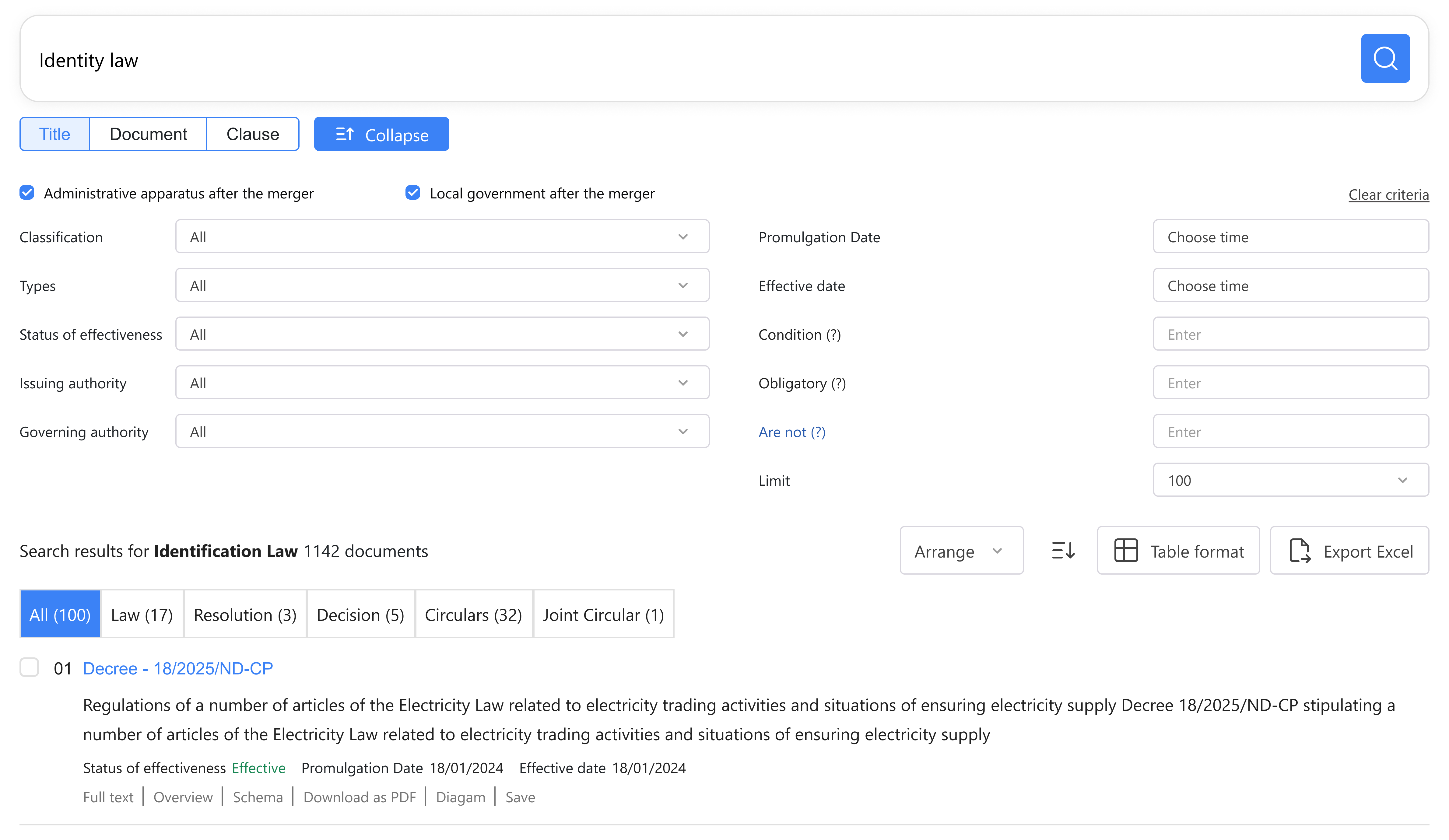}
    \caption{Legal document retrieval tool: we provide a search interface that enables legal experts to efficiently locate relevant legal documents and supporting materials for the annotation process.}
    \label{fig:search_tool}
\end{figure*}

After constructing the Legal Corpus Database and the Knowledge Graph Database to store currently effective legal documents, covering a wide range of legal topics and real-world citizen questions submitted to law offices, we developed search and retrieval tools to support legal experts in efficiently locating relevant Articles, Clauses, Points, and legal topics. These tools are used to create task-specific raw data and to annotate samples into either multiple-choice or generative question formats for each benchmark task. Illustrations of the tools are shown in Figure \ref{fig:annotation_tool} and Figure \ref{fig:search_tool}.
\section{Details of Contamination Study}\label{appen:contamination}
To assess data contamination risk, we conducted a multi-pronged analysis on 1000 stratified instances covering all task categories using complementary detection methods. We randomly sampled up to 20 13-grams from each input question (excluding instructions, options, and labels) and performed exact-match Google searches, flagging instances as potentially contaminated if at least two distinct 13-grams matched the same webpage within the top-5 results beyond standard statutory references. To address potential under-indexing of Vietnamese legal content, we additionally conducted targeted searches on major Vietnamese legal portals (e.g., \texttt{thuvienphapluat.vn}, \texttt{chinhphu.vn}) using both exact 13-gram matching and fuzzy matching with a similarity threshold of 0.85. We further queried representative samples against Common Crawl indices (2020-2024) using exact n-gram matching and semantic similarity search. To distinguish genuine contamination from ubiquitous legal language, all matches were categorized into mandatory statutory text, standard legal templates, and substantive case content, with only the latter counted as true contamination. Across all methods, only 1.8\% of instances exhibited potential overlap, all attributable to statutory provisions or standardized templates by design; no overlap was found for case-based, reasoning, or generation tasks, and no additional contamination was detected beyond Google search results. Overall, these findings indicate minimal contamination risk and confirm that VietLegal primarily evaluates legal reasoning rather than memorization.
\section{Details of Tested LLMs}\label{appen:model}
Table~\ref{tab:models} summarizes detailed information about the LLMs evaluated on VietLegal.
\begin{table*}[t]
\centering
\small
\setlength{\tabcolsep}{4pt}
\caption{Large language models evaluated on VietLegal.}
\label{tab:models}
\begin{tabular}{llccp{5.5cm}}
\toprule
\textbf{Model Type} & \textbf{Model} & \textbf{Size} & \textbf{Context} & \textbf{Access \& URL} \\
\midrule
\multirow{14}{*}{\parbox{2.2cm}{\centering\textbf{General}\\\textbf{Multilingual}\\\textbf{LLMs}}}
& GPT-4o & - & 128k & API: \url{platform.openai.com} \\
& GPT-4o-mini & - & 128k & API: \url{platform.openai.com} \\
& Claude 4.5 Sonnet & - & 200k & API: \url{platform.claude.com} \\
& Gemini 2.5 Flash & - & 1M & API: \url{ai.google.dev} \\
& Qwen 2.5 Instruct & 3B & 32k & Weights: \url{huggingface.co/Qwen} \\
& Qwen 2.5 Instruct & 7B $\rightarrow$ 72B & 128k & Weights: \url{huggingface.co/Qwen} \\

& Llama 2 Chat & 7B/13B & 4k & Weights: \url{huggingface.co/meta-llama} \\
& Llama 3 Instruct & 8B/70B & 8k & Weights: \url{huggingface.co/meta-llama} \\
& InternLM 3 Instruct & 8B & 8k & Weights: \url{huggingface.co/internlm} \\
& InternLM Chat & 20B & 16k & Weights: \url{huggingface.co/internlm} \\
& Gemma 2 Instruct & 9B/27B & 8k & Weights: \url{huggingface.co/google} \\
\midrule
\multirow{6}{*}{\parbox{1.8cm}{\centering\textbf{Vietnamese-}\\\textbf{focused LLMs}}}
& SeaLLMs v3 Chat & 1.5B & 8k & Weights: \url{huggingface.co/SeaLLMs} \\
& SeaLLMs v3 Chat & 7B & 8k & Weights: \url{huggingface.co/SeaLLMs} \\
& BloomVN Chat & 8B & 8k & Weights: \url{huggingface.co/BlossomsAI} \\
& \parbox{2cm}{Qwen 3 4B\\Legal Pretrain} & 4B & 8k & Weights: \url{hf.co/VLSP2025-LegalSML} \\
& CMC-Legal & 32B & 128k & Weights: \url{huggingface.co/CMC-OPENAI} \\
\bottomrule
\end{tabular}
\end{table*}

\subsection{CMC-Legal-32B}\label{appen:proprietary_legal}
 
CMC-Legal-32B is a domain-adapted version of Qwen2.5-32B-Instruct. It keeps the base model's architecture, parameter count, tokenizer, vocabulary, and 128k context length; we evaluate the checkpoint released in January 2026. Because it shares architecture and size with Qwen2.5-32B-Instruct, the comparison between the two controls for these factors. However, the adaptation combines Vietnamese-language and legal-domain training, so performance differences should be attributed to the overall Vietnamese legal adaptation pipeline rather than to legal data alone.
 
\paragraph{Stage 1: continued pretraining.} The model was further pretrained on 79.2\,GB of Vietnamese text, of which 23\,GB are curated legal documents collected from official government sources (the national legal database, ministry portals, published judgments, and precedents), comprising 568{,}910 legal documents across 42 document types.
 
\paragraph{Stage 2: supervised fine-tuning.} The model was then fine-tuned on approximately 198{,}800 legal instruction-response pairs covering legal retrieval, question answering, summarisation, and instruction following.
 
\paragraph{Decontamination.} Since VietLegal was constructed before this training, all training data were decontaminated against the benchmark using the protocol of Appendix~\ref{appen:contamination}.

\section{Details of Labelling Process}\label{appen:labelling}
\textbf{Recruitment.} Our annotation team comprised three senior legal experts and eight junior legal experts, all recruited through partnerships with two Vietnamese law firms and one university law faculty. Senior experts (Teachers) were required to hold a valid Vietnamese lawyer's license with a minimum of five years of professional practice, specialization in at least one of the benchmark's core domains (civil law, criminal law, administrative law, or commercial law), and prior experience in legal education or training. Junior experts were licensed lawyers or final-year law graduates who had passed the Vietnamese bar examination, with 1-3 years of practical experience in legal research, case preparation, or client consultation. All annotators were native Vietnamese speakers.

Before annotation, all team members completed a two-day training program consisting of: (1) an overview of the benchmark's cognitive framework and task definitions, (2) hands-on practice sessions using a pilot set of 50 samples per task with immediate feedback, and (3) calibration exercises where annotators discussed edge cases and established shared labeling conventions. Training materials included detailed annotation guidelines specifying decision rules for ambiguous cases, such as how to handle repealed-but-referenced articles or provisions with multiple valid interpretations. Annotators were required to achieve at least 85\% agreement with gold-standard pilot labels before proceeding to the main annotation.
Each annotator was compensated at a rate of 150,000 VND (approximately 6 USD) per hour, consistent with professional legal consultation rates in Vietnam. Senior experts received an additional supervision stipend. The total annotation effort spanned approximately 1,400 person-hours over 14 weeks. To mitigate fatigue effects, annotators were limited to 4-hour sessions with mandatory breaks, and task assignments were rotated weekly to prevent over-specialization. All annotators provided informed consent for their contributions to be used in academic research.

\textbf{Labelling Process.}
Using the two constructed databases, we design benchmark questions spanning 22 tasks, aligned with predefined cognitive levels and target sample sizes (Table~\ref{tab:benchmark_tasks}). Annotation follows a structured multi-stage expert-in-the-loop protocol. A senior legal expert (licensed lawyer with over 5 years of professional experience across civil, criminal, and administrative law) supervises the process by defining task-specific topics and identifying authoritative legal sources. Relevant documents are retrieved, and task-specific raw data are prepared accordingly. For each task, the sample quota is evenly split between two independent junior legal experts (lawyers with 1-3 years of experience), denoted as Junior A and Junior B, who independently construct realistic legal scenarios and corresponding answers in either multiple-choice or open-ended formats. To ensure quality, a batch-wise cross-verification procedure is applied: after every \textbf{100 samples}, Junior A and Junior B exchange batches and independently answer each other's questions in a blind setting.

\textbf{Verifying Process.}
Inter-annotator agreement is evaluated using percentage agreement and Cohen’s Kappa, measured on identical labels assigned \emph{before any discussion}. Across 10{,}450 samples, initial agreement reaches \textbf{92.39\%} (\textbf{9{,}656}/10{,}450) with a Cohen’s Kappa of \textbf{0.89}, indicating strong consistency beyond chance. For the remaining \textbf{7.61\%} (794/10{,}450) disputed samples, a two-stage resolution process is applied. First, the two junior annotators conduct structured discussions and re-examine legal sources, resolving 683 cases by consensus. The remaining 111 cases are escalated to the senior legal expert for final adjudication. These unresolved cases are concentrated in high-complexity tasks, notably Conflict and Consistency Detection (Task~3.4, 31 cases), Multi-Article Reasoning (Task~3.3, 27 cases), and Unfair Contract Detection (Task~5.4, 22 cases), which require multi-provision synthesis, conflict resolution, and nuanced normative judgment. During annotation, dedicated retrieval tools are provided to support efficient search and verification over the constructed databases. The final benchmark contains 10{,}450 expert-validated legal instances, each explicitly grounded in authoritative legal sources. Details of the annotation support tool are provided in Appendix~\ref{appen:annotate}.

\section{Transferability of the Design Principle to Other Civil-Law Systems}\label{appen:transferability}
 
This section delineates the scope of VietLegal's generalizability. We do not posit that the VietLegal dataset can be directly transferred to other jurisdictions. Rather, our claim concerns the transferability of the underlying \emph{cognitive evaluation framework} to civil-law systems characterised by hierarchical and codified statutory structures. In contrast, the \emph{benchmark instantiation}, including statutes, legal terminology, case law, and professional rules, remains jurisdiction-specific and must be adapted to the target legal system.
 
\subsection{Why the Framework Transfers}
 
Vietnam has a socialist legal system that is fundamentally shaped by the civil-law tradition~\cite{merryman2018civil}. Like representative civil-law jurisdictions such as France and Germany, whose civil codes strongly influenced the development of Vietnamese private law, as well as Japan, South Korea, and China, it shares two characteristics that motivate our taxonomy.
 
\paragraph{Statutory law is the primary source of law (Levels 1--2).} Law is organized as hierarchical bodies of legislation spanning multiple legal domains, with clear normative hierarchies and frequent amendment and consolidation. These characteristics motivate Levels~1--2, which evaluate retrieving legal sources, understanding statutory structure, tracing cross-references, and organizing legal information.
 
\paragraph{Reasoning proceeds from rule to facts (Levels 3--5).} Legal reasoning starts by identifying the applicable statutory rule and then applies it to concrete facts, rather than creating new rules through precedent. The internal structure of statutory rules (conditions, legal consequences, and sanctions) gives rise to a reasoning process shared across civil-law jurisdictions: identifying the governing provision, interpreting its requirements, applying it to the facts, and articulating the resulting legal analysis. This progression motivates Levels~3--5, which evaluate rule application, legal reasoning, and higher-order professional judgment under ethical and fairness constraints.
 
VietLegal was therefore designed to reflect the cognitive workflow of legal practitioners in codified systems, not the substantive content of Vietnamese law. Statutory provisions, legal terminology, case materials, procedural rules, and professional-conduct regulations must be adapted to each jurisdiction, while the underlying cognitive operations stay the same.
 
\subsection{Mapping to Existing Civil-Law Benchmarks}
 
Table~\ref{tab:transferability} maps each VietLegal level to its closest analogue in existing civil-law benchmarks: LawBench~\cite{fei2024lawbench} (Chinese), BSARD~\cite{louis2022french} (French/Belgian), and GerLayQA~\cite{buttner2024german} (German). Levels~1--4 all have counterparts, which indicates that the cognitive operations they target are already recognized as meaningful outside Vietnam. Level~5 has no direct counterpart and is a distinctive contribution of VietLegal.
 
\begin{table*}[h]
\centering
\small
\renewcommand{\arraystretch}{1.25}
\caption{Mapping VietLegal's cognitive levels to existing civil-law benchmarks, and what must change to transfer each level to a new jurisdiction.}
\label{tab:transferability}
\begin{tabular}{p{0.17\textwidth}p{0.42\textwidth}p{0.33\textwidth}}
\toprule
\textbf{VietLegal level} & \textbf{Existing civil-law benchmark analogue} & \textbf{Transferability} \\
\midrule
Level 1 -- Recognition \& Recall &
LawBench: legal knowledge memorization (article recitation, legal knowledge QA) &
Directly transferable after replacing the statutory corpus with that of the target jurisdiction. \\
\midrule
Level 2 -- Understanding \& Structuring &
LawBench: legal knowledge understanding (NER, event detection, relation extraction); BSARD: statutory structure and provision organization &
Directly transferable, as these tasks operate on the hierarchical organization of statutory law. \\
\midrule
Level 3 -- Reasoning \& Inference &
LawBench: article, charge, and prison-term prediction, case analysis; BSARD and GerLayQA: statutory retrieval and grounded question answering &
Directly transferable, since every civil-law system requires applying statutory rules to facts; jurisdiction-specific rules must be substituted. \\
\midrule
Level 4 -- Interpretation \& Generation &
LawBench: legal consultation and summarization; GerLayQA: grounded legal answer generation &
Transferable at the cognitive level; the legal-writing format (e.g., IRAC or the German \emph{Gutachtenstil}) should follow local practice. \\
\midrule
Level 5 -- Ethics, Fairness \& Bias &
No direct counterpart in existing benchmarks &
Conceptually transferable, but evaluation criteria depend on each jurisdiction's professional ethics, privacy regulations, and legal standards. \\
\bottomrule
\end{tabular}
\end{table*}
 
\subsection{Components that Require Redesign}
 
Some components must be redesigned rather than transferred directly. Task~2.3 (Legal Knowledge Graph Structuring) encodes the Article--Clause--Point hierarchy of Vietnamese legislation, so it must be adapted to the document hierarchy of the target jurisdiction (e.g., \emph{Buch}--\emph{Abschnitt}--\emph{Paragraph} in German codes). Likewise, Level~5 must be instantiated with the target jurisdiction's professional-conduct rules, privacy regulations, and ethical standards. These adaptations affect the benchmark \emph{content}, not the cognitive framework itself.

\section{More Experimental Results}\label{appen:more_result}
\subsection{Human Evaluation Protocol}

To assess the quality of model-generated legal responses, we conducted a rigorous human evaluation on generation tasks (Tasks~4.1-4.3). We randomly sampled 20 instances per task per model, resulting in 300 total samples (20 instances $\times$ 3 tasks $\times$ 5 systems, including human ground truth). 

\paragraph{Annotation Procedure} Two junior legal experts independently evaluated all samples. To ensure unbiased assessment, we employed a double-blind protocol: (1) model-generated and human-written responses were randomly shuffled and presented without any identifying information, and (2) annotators were not informed which responses were human-authored versus machine-generated. Each sample was evaluated using Senior expert-defined criteria on a 1-5 Likert scale: \emph{Legal Accuracy}, measuring the correctness and appropriateness of legal reasoning and statutory references, and \emph{Completeness}, measuring whether the response sufficiently addressed all legally relevant aspects of the query.

\paragraph{Quality Control} To verify annotation consistency, we computed Cohen's kappa for inter-annotator agreement across all rated dimensions. The mean $\kappa = 0.92$ (substantial agreement) confirms high reliability in expert judgments. Final scores were computed as the mean of both annotators' ratings. The total annotation effort required approximately 40 hours across both evaluators. To validate blinding effectiveness, annotators were asked post-hoc whether they could identify human versus model responses; both reported uncertainty, confirming successful blinding.

\begin{table*}[t]
\centering
\scriptsize
\caption{Human evaluation of generated answers and ground-truth responses on interpretation and generation tasks (Tasks~4.1-4.3). Two legal experts independently rated each sample based on \emph{Legal Accuracy} and \emph{Completeness} using a 1-5 Likert scale. Evaluators were blinded to the source of each response (human-written or model-generated).}
\label{tab:human-eval}
\begin{tabular}{lccccccc}
\toprule
\multirow{2}{*}{\textbf{Model}} & \multicolumn{2}{c}{\textbf{Task 4.1}} & \multicolumn{2}{c}{\textbf{Task 4.2}} & \multicolumn{2}{c}{\textbf{Task 4.3}} \\
\cmidrule(lr){2-3} \cmidrule(lr){4-5} \cmidrule(lr){6-7}
& Legal Accuracy & Completeness & Legal Accuracy & Completeness & Legal Accuracy & Completeness \\
\midrule
gpt-4o & 3.59 & 3.18 & 2.62 & 2.59 & 3.27 & 3.26 \\
gpt-4o-mini & 3.39 & 3.01 & 2.36 & 2.36 & 3.30 & 3.38 \\
Qwen2.5-72B-Instruct & 3.34 & 2.76 & 2.40 & 2.53 & 3.23 & 3.16 \\
CMC-Legal-32B & 3.79 & 3.38 & 2.53 & 2.62 & 3.57 & 3.55 \\
Human & 4.79 & 4.88 & 4.76 & 4.78 & 4.82 & 4.85 \\
\bottomrule
\end{tabular}
\end{table*}

Table~\ref{tab:human-eval} reports human evaluation results across interpretation and generation tasks. Across all tasks, human-written responses achieve substantially higher scores (Legal Accuracy: 4.76-4.82; Completeness: 4.78-4.88), establishing an upper bound for current systems. Among models, the domain-adapted CMC-Legal-32B consistently outperforms general-purpose LLMs on both metrics, with particularly notable advantages on Tasks~4.1 and~4.3 (Legal Accuracy: 3.79 and 3.57 respectively). This pattern suggests that legal-domain pretraining and fine-tuning improve factual correctness and coverage in interpretive legal writing. GPT-4o demonstrates competitive performance, achieving the second-highest scores on most tasks, though it lags behind CMC-Legal-32B on legal accuracy metrics.

Nevertheless, a substantial gap remains between model-generated outputs and expert-authored responses across all tasks and metrics (average gap: 1.2-1.5 points on the 5-point scale), with the disparity most pronounced in completeness scores. This gap is especially evident in Task~4.2, where all models struggle (Legal Accuracy: 2.36-2.62; Completeness: 2.36-2.62), suggesting particular difficulty with this task type. These results highlight persistent limitations in current LLMs' ability to produce fully grounded and exhaustive legal interpretations, even when surface-level fluency appears strong. The high inter-annotator agreement ($\kappa = 0.92$) provides confidence that these quality differences reflect genuine performance gaps rather than annotation noise.

\paragraph{How well do automatic metrics track legal correctness?}
We compute the Spearman correlation between each automatic metric and expert-rated Legal Accuracy. Since per-model samples are small ($n=20$), we compute the correlation within each of the four models and combine them with Fisher's $z$-transformation (effective $n=80$ per task), which avoids inflated correlations from pooling scores across models. Table~\ref{tab:metric_correlation} shows moderate positive correlations throughout ($\rho = 0.44$--$0.66$): the metrics are useful scalable proxies but do not substitute for expert judgment. BERTScore aligns better than ROUGE-L on the two reasoning-intensive tasks (4.2 and 4.3), suggesting that semantic similarity captures legal correctness better than lexical overlap when the model must produce reasoning rather than a summary. Accordingly, all conclusions in this paper about the legal correctness of generated text rest on the expert evaluation, and automatic scores are used only for relative comparison.
 
\begin{table}[t]
\centering
\scriptsize
\caption{Spearman correlation between automatic metrics and expert-rated Legal Accuracy, computed within each of four models and combined via Fisher's $z$ (effective $n=80$ per task).}
\label{tab:metric_correlation}
\begin{tabular}{llccc}
\toprule
\textbf{Task} & \textbf{Metric} & $\boldsymbol{\rho}$ & \textbf{95\% CI} & $\boldsymbol{p}$ \\
\midrule
\multirow{2}{*}{4.1 Summarization} & ROUGE-L   & 0.66 & [0.51, 0.78] & $1.4\times10^{-10}$ \\
                                   & BERTScore & 0.65 & [0.49, 0.77] & $5.2\times10^{-10}$ \\
\midrule
\multirow{2}{*}{4.2 Judicial Reasoning} & ROUGE-L   & 0.44 & [0.23, 0.62] & $1.4\times10^{-4}$ \\
                                        & BERTScore & 0.56 & [0.37, 0.71] & $3.9\times10^{-7}$ \\
\midrule
\multirow{2}{*}{4.3 Legal Opinion} & ROUGE-L   & 0.52 & [0.32, 0.67] & $4.2\times10^{-6}$ \\
                                   & BERTScore & 0.56 & [0.38, 0.71] & $3.0\times10^{-7}$ \\
\bottomrule
\end{tabular}
\end{table}

\begin{table*}[t]
\centering
\scriptsize
\caption{Zero-shot performance with reasoning on Understanding \& Structuring (Level 2) and Reasoning \& Inference (Level 3) tasks. R-L denotes ROUGE-L score. For Task 3.4 (Conflict Detection), we report Y-F1/N-F1 scores. \textcolor{red}{\textbf{Red}} indicates best overall results and \textcolor{blue}{\textbf{blue}} indicates best open source results in each column.}
\label{tab:reasoning_results}
\begin{tabular}{l|ccccc|ccccc}
\toprule
\multirow{2}{*}{\textbf{Model}} & \multicolumn{5}{c|}{\textbf{Understanding \& Structuring }} & \multicolumn{5}{c}{\textbf{Reasoning \& Inference }} \\
\cmidrule(lr){2-6} \cmidrule(lr){7-11}
& \textbf{2.1} & \textbf{2.2} & \textbf{2.3} & \textbf{2.4} & \textbf{2.5} & \textbf{3.1} & \textbf{3.2} & \textbf{3.3} & \textbf{3.4} & \textbf{3.5} \\
& Acc & Acc & R-L & Acc & m-F1 & Acc & Acc & Acc & Y-F1/N-F1 & Acc \\
\midrule
gpt-4o & 75.86 & 55.77 & \textcolor{red}{\textbf{0.630}} & 79.97 & 55.29 & 39.00 & 81.36 & \textcolor{red}{\textbf{77.93}} & 2.35/41.96 & \textcolor{red}{\textbf{66.48}} \\
gpt-4o-mini & 75.86 & 50.16 & 0.612 & 79.30 & 55.00 & 34.67 & 78.67 & 72.26 & 1.59/39.81 & 53.48 \\
Gemini-2.5-flash & \textcolor{red}{\textbf{78.54}} & \textcolor{red}{\textbf{67.35}} & 0.456 & 73.08 & 55.95 & \textcolor{red}{\textbf{50.60}} & 84.81 & 75.37 & 17.46/42.86 & 66.01 \\
\midrule
Qwen2.5-32B-Instruct & \textcolor{blue}{\textbf{73.72}} & 62.33 & \textcolor{blue}{\textbf{0.612}} & \textcolor{red}{\textbf{80.63}} & 59.78 & 32.67 & 84.00 & 71.23 & 0.00/39.61 & 54.32 \\
Qwen2.5-72B-Instruct & 74.12 & \textcolor{blue}{\textbf{66.33}} & 0.449 & 79.97 & \textcolor{red}{\textbf{60.20}} & 35.50 & 81.34 & \textcolor{blue}{\textbf{75.00}} & 9.16/40.80 & \textcolor{blue}{\textbf{58.77}} \\
Qwen2.5-14B-Instruct & 74.90 & 63.33 & 0.577 & 77.13 & 59.76 & 35.17 & 80.83 & 70.55 & 0.00/39.61 & 53.48 \\
Qwen2.5-7B-Instruct & 73.33 & 53.67 & 0.376 & 73.12 & 46.93 & \textcolor{blue}{\textbf{40.33}} & 78.67 & 64.73 & 3.15/40.00 & 49.86 \\
Qwen2.5-3B-Instruct & 60.00 & 51.67 & 0.171 & 67.45 & 53.41 & 27.33 & 66.33 & 49.66 & 4.62/38.61 & 45.68 \\
Llama-3.1-8B-Instruct & 56.92 & 37.67 & 0.267 & 64.77 & 54.32 & 31.17 & 70.00 & 60.27 & 1.59/39.81 & 53.20 \\
Llama-3.1-70B-Instruct & 70.98 & 60.67 & 0.429 & 72.78 & 57.79 & 37.33 & 81.67 & 73.63 & 1.59/39.82 & 61.56 \\
CMC-Legal-32B & 70.43 & 64.00 & 0.514 & 79.80 & 59.20 & 38.17 & \textcolor{red}{\textbf{85.56}} & 64.84 & \textcolor{red}{\textbf{34.00}}/\textcolor{red}{\textbf{56.00}} & 56.82 \\
\bottomrule
\end{tabular}
\end{table*}

\begin{table*}[t]
\centering
\scriptsize
\setlength{\tabcolsep}{3pt}
\caption{Few-shot performance without reasoning on Understanding \& Structuring (Level 2), Reasoning \& Inference (Level 3), and Ethics, Fairness \& Bias (Level 5) tasks. R-L denotes ROUGE-L score. For Task 3.4 (Conflict Detection), we report Y-F1/N-F1 scores. \textcolor{red}{\textbf{Red}} indicates best overall results and \textcolor{blue}{\textbf{blue}} indicates best open-source results in each column.}
\label{tab:fewshot_no_reasoning}
\begin{tabular}{l|ccccc|ccccc|cccc}
\toprule
\multirow{2}{*}{\textbf{Model}} & \multicolumn{5}{c|}{\textbf{Understanding \& Structuring}} & \multicolumn{5}{c|}{\textbf{Reasoning \& Inference}} & \multicolumn{4}{c}{\textbf{Ethics, Fairness \& Bias}} \\
\cmidrule(lr){2-6} \cmidrule(lr){7-11} \cmidrule(lr){12-15}
& \textbf{2.1} & \textbf{2.2} & \textbf{2.3} & \textbf{2.4} & \textbf{2.5} & \textbf{3.1} & \textbf{3.2} & \textbf{3.3} & \textbf{3.4} & \textbf{3.5} & \textbf{5.1} & \textbf{5.2} & \textbf{5.3} & \textbf{5.4} \\
& Acc & Acc & R-L & Acc & m-F1 & Acc & Acc & Acc & Y-F1/N-F1 & Acc & Acc & Acc & Acc & Acc \\
\midrule
gpt-4o & \textcolor{red}{\textbf{86.16}} & 68.33 & 0.657 & 85.64 & 56.83 & 40.67 & 85.33 & \textcolor{red}{\textbf{78.77}} & 77.67/62.30 & \textcolor{red}{\textbf{69.64}} & 43.37 & \textcolor{red}{\textbf{69.12}} & 92.04 & 63.68 \\
gpt-4o-mini & 78.27 & 58.33 & 0.623 & 85.31 & 61.83 & 39.67 & 82.50 & 77.05 & 44.58/44.58 & 59.05 & 44.18 & 65.44 & 87.55 & 46.15 \\
Gemini-2.5-flash & 78.66 & 61.67 & 0.422 & 83.81 & 53.18 & \textcolor{red}{\textbf{50.17}} & 86.00 & 76.37 & \textcolor{red}{\textbf{85.71}}/0.00 & 65.18 & 46.98 & 66.36 & 91.05 & 65.38 \\
\midrule
Qwen2.5-32B-Instruct & 69.80 & 67.00 & 0.755 & 86.81 & 57.59 & 39.67 & 84.83 & 75.34 & \textcolor{blue}{42.00}/35.44 & 59.05 & \textcolor{red}{\textbf{56.59}} & \textcolor{blue}{\textbf{68.67}} & 93.04 & 70.51 \\
Qwen2.5-72B-Instruct & 72.16 & 65.67 & 0.557 & \textcolor{red}{\textbf{88.98}} & \textcolor{blue}{\textbf{69.00}} & 38.50 & 86.17 & 75.34 & 0.00/39.60 & 63.79 & 47.00 & 66.82 & 91.04 & 62.39 \\
Qwen2.5-14B-Instruct & 75.89 & \textcolor{red}{\textbf{69.67}} & 0.716 & 84.47 & 59.27 & 43.33 & 81.67 & \textcolor{blue}{\textbf{75.00}} & 0.00/39.61 & 56.55 & 47.00 & \textcolor{blue}{\textbf{68.67}} & 93.02 & 66.67 \\
Qwen2.5-7B-Instruct & 77.65 & 57.00 & 0.589 & 83.31 & 59.98 & 39.50 & 80.17 & 70.21 & 0.00/39.62 & 58.22 & 38.15 & 60.37 & 91.03 & 65.81 \\
Qwen2.5-3B-Instruct & 64.82 & 57.33 & 0.693 & 71.45 & 57.71 & 34.17 & 74.33 & 68.49 & 4.62/38.61 & 51.33 & 36.55 & 52.53 & 85.98 & 61.54 \\
Llama-3.1-8B-Instruct & 76.28 & 34.67 & 0.738 & 64.94 & 44.35 & 29.00 & 78.67 & 34.93 & 1.59/39.81 & 43.18 & 37.35 & 48.39 & 60.56 & 62.39 \\
Llama-3.1-70B-Instruct & 76.68 & 68.33 & 0.693 & 82.30 & 61.68 & \textcolor{red}{\textbf{45.83}} & 85.50 & 77.05 & 0.00/39.61 & \textcolor{blue}{64.07} & 41.77 & 65.44 & 93.03 & 65.81 \\
CMC-Legal-32B & \textcolor{blue}{\textbf{82.61}} & 58.67 & \textcolor{red}{\textbf{0.791}} & 88.48 & 60.05 & 45.50 & \textcolor{red}{\textbf{90.50}} & \textcolor{blue}{\textbf{75.00}} & 42.00/\textcolor{red}{\textbf{75.36}} & 63.51 & 36.55 & 64.06 & \textcolor{red}{\textbf{94.52}} & \textcolor{red}{\textbf{80.77}} \\
\bottomrule
\end{tabular}
\end{table*}

% We conduct systematic ablation studies to disentangle the effects of \textbf{explicit reasoning} (chain-of-thought prompting) and \textbf{task demonstrations} (few-shot prompting) under controlled settings. Specifically, we compare:
% (i) \textbf{Zero-shot without reasoning} (baseline, reported in main results),
% (ii) \textbf{Zero-shot with reasoning}, and
% (iii) \textbf{Few-shot without reasoning}.
% Our analysis focuses on tasks where prompting strategies are most consequential: Understanding \& Structuring (Level~2), Reasoning \& Inference (Level~3), and Ethics, Fairness \& Bias (Level~5).

\subsection{Zero-shot with Reasoning}
\label{subsec:zeroshot_reasoning}

Table~\ref{tab:reasoning_results} reports zero-shot performance when models are explicitly instructed to generate intermediate reasoning steps.

\paragraph{Understanding \& Structuring (Level 2).}
Contrary to common assumptions, explicit reasoning does not reliably improve and often degrades performance on document understanding and structured extraction tasks. While some models show modest gains on holistic comprehension tasks (e.g., Tasks~2.1 and~2.2), many experience noticeable drops, particularly on structurally grounded tasks such as legal information extraction and alignment (Task~2.3, ROUGE-L). For instance, large general-purpose models (e.g., GPT-4o, Qwen2.5-72B) suffer consistent declines when forced to articulate step-by-step reasoning.

This pattern suggests a \textit{reasoning paradox}: Level~2 legal tasks rely heavily on holistic pattern recognition over complex legal texts rather than explicit logical decomposition. Forcing intermediate reasoning introduces cognitive overhead that interferes with direct matching between legal schemas and surface forms, indicating that explicit chain-of-thought is poorly aligned with the nature of structured legal understanding.

\paragraph{Reasoning \& Inference (Level 3).}
In contrast, explicit reasoning provides clearer but task-dependent benefits for inference-heavy problems. Procedural and multi-step tasks (e.g., Tasks~3.1 and~3.3) often improve under reasoning prompts, with models such as Gemini-2.5-Flash achieving the highest accuracy on article prediction. However, gains are inconsistent: conflict detection (Task~3.4) remains unstable, with many models collapsing to a single label despite reasoning instructions. Notably, the domain-adapted CMC-Legal-32B model achieves the most balanced Yes/No F1 scores, highlighting that domain-specific representations are more effective than generic reasoning prompts for nuanced legal judgments.

Overall, zero-shot reasoning selectively benefits tasks requiring explicit inference chains, while proving ineffective or even harmful for structurally grounded legal understanding.

\subsection{Few-shot without Reasoning}
\label{subsec:fewshot_noreasoning}

Table~\ref{tab:fewshot_no_reasoning} presents results under few-shot prompting without explicit reasoning instructions.

\paragraph{Understanding \& Structuring (Level 2).}
Few-shot demonstrations lead to substantial and consistent improvements across all models, often exceeding the gains from zero-shot reasoning. Example-based prompting enables models to internalize task-specific formats and legal schemas, resulting in strong performance even without articulated reasoning. Several open-source models approach or surpass proprietary models under zero-shot reasoning, and CMC-Legal-32B achieves the highest ROUGE-L score on Task~2.3, underscoring the effectiveness of demonstrations for legal text restructuring.

\paragraph{Reasoning \& Inference (Level 3).}
Few-shot prompting frequently matches or outperforms zero-shot reasoning, particularly on Tasks~3.1 and~3.5. These results suggest that demonstrations implicitly induce reasoning behaviors, allowing models to infer decision patterns without explicit chain-of-thought outputs.

\textbf{Conflict Detection (Task~3.4)} is especially revealing. In zero-shot settings, most models exhibit extreme label imbalance, often predicting ``no conflict'' exclusively. Few-shot demonstrations fundamentally alter this behavior: models such as GPT-4o and Gemini-2.5-Flash transition from near-total failure to balanced and competent conflict detection. This indicates that abstract legal notions like inconsistency or contradiction cannot be reliably conveyed through instructions alone; concrete examples are essential for grounding such concepts in context.

\paragraph{Ethics, Fairness \& Bias (Level 5).}
Few-shot prompting yields more stable and robust performance on ethical judgment tasks. Domain-adapted models, particularly CMC-Legal-32B, benefit substantially from demonstrations, achieving strong alignment with Vietnamese legal norms on tasks such as unfair contract detection (Task~5.4). Interestingly, some mid-scale open-source models outperform larger proprietary models under few-shot settings, suggesting that ethical judgments are especially sensitive to contextual grounding rather than raw model scale.

\begin{table*}[h]
\centering
\scriptsize
\caption{Ablation study comparing Zero-shot No Reasoning and Agentic RAG across reasoning tasks. Blue indicates improvement, red indicates degradation. Numbers in parentheses show the performance delta.}
\label{tab:ablation_rag_results}
\begin{tabular}{l|c|cccc}
\hline
\textbf{Model} & \textbf{Setting} & \textbf{3.1 Acc} & \textbf{3.2 Acc} & \textbf{3.3 Acc} & \textbf{3.4 Y-F1/N-F1} \\
\hline
\multirow{2}{*}{GPT-4o} 
& Zero-shot No Reasoning & 38.83 & 84.50 & 73.50 & 27.21/42.16 \\
& Agentic RAG & 60.83 (\textcolor{blue}{+22.00}) & 82.00 (\textcolor{red}{-2.50}) & 72.9 (\textcolor{red}{-0.60}) & 53.15/40.00 \\
\hline
\multirow{2}{*}{GPT-4o-mini} 
& Zero-shot No Reasoning & 35.33 & 82.17 & 74.66 & 11.85/39.59 \\
& Agentic RAG & 54.17 (\textcolor{blue}{+18.84}) & 80.83 (\textcolor{red}{-1.34}) & 70.89 (\textcolor{red}{-3.77}) & 94.65/39.41 \\
\hline
\multirow{2}{*}{Gemini-2.5-Flash} 
& Zero-shot No Reasoning & 40.83 & 84.67 & 76.71 & 0.00/39.61 \\
& Agentic RAG & 53.05 (\textcolor{blue}{+12.22}) & 81.62 (\textcolor{red}{-3.05}) & 71.58 (\textcolor{red}{-5.13}) & 87.69/40.59 \\
\hline
\multirow{2}{*}{CMC-Legal-32B} 
& Zero-shot No Reasoning & 41.67 & 90.67 & 76.71 & 86.41/13.33 \\
& Agentic RAG & 52.33 (\textcolor{blue}{+10.66}) & 88.67 (\textcolor{red}{-2.00}) & 74.31 (\textcolor{red}{-2.40}) & 74.00/60.61 \\
\hline
\end{tabular}
\end{table*}

\subsection{Retrieval-Augmented Generation setting}
\label{sec:ablation_rag}

To evaluate the impact of retrieval on legal reasoning, we conduct an ablation study comparing two settings. The first, \textbf{Zero-shot Reasoning}, is identical to the evaluation setup used for the main results in the paper, serving as the primary non-retrieval baseline. The second, \textbf{Agentic Retrieval-Augmented Generation} \cite{singh2025agentic}, extends this setting by enabling models to access an external legal corpus through a search tool.

\paragraph{Zero-shot Reasoning (Baseline).}
In this setting, models are provided with only the task instruction and the question with answer options (see Table~\ref{tab:legal_reference_retrieval_example} for an example). No additional legal documents or contextual information are included in the prompt. Models must rely solely on their internal knowledge acquired during pre-training to answer questions. This represents a pure zero-shot evaluation where models cannot access external information.

\paragraph{Agentic RAG.}
Models are equipped with a search tool that provides access to our Legal Corpus Database containing over 55,000 statutory documents. Each document is segmented into chunks of 1024 tokens with 100-token overlap to preserve context across boundaries. We implement a hybrid retrieval system combining three complementary approaches: (1) \textit{dense retrieval} using Vietnamese legal-domain fine-tuned PhoBERT embeddings with FAISS indexing for semantic matching; (2) \textit{sparse retrieval} using BM25 with legal-specific tokenization that preserves article numbers and legal references for keyword-based matching; and (3) \textit{exact-match search} via Elasticsearch for precise statute and phrase matching. The final ranking employs Reciprocal Rank Fusion (RRF) with weights (0.4 dense, 0.3 sparse, 0.3 exact) optimized on a validation set, retrieving the top-5 chunks per query. Unlike standard RAG with single-shot retrieval, our agentic approach allows models to iteratively interact with the database: they generate natural language search queries, retrieve relevant chunks with metadata (document title, article/clause numbers, enactment date, amendment status), analyze results, and reformulate queries or issue additional searches as needed, with a maximum of 5 iterations. To handle legal document versioning, we implement temporal filtering that prioritizes documents valid at the query's reference time and provides amendment chain information. 

\subsubsection{Results and Analysis}

Table~\ref{tab:ablation_rag_results} reports performance across all reasoning tasks.

\paragraph{Task 3.1 (Article/Clause Identification).}
Agentic RAG yields substantial improvements across all models (ranging from +10.66 to +22.00 points), with the largest gain observed for GPT-4o. This task requires locating specific statutory provisions, which directly benefit from tool-based retrieval over a large legal corpus. Retrieved chunks typically contain complete articles with sufficient surrounding context for accurate clause identification.

In contrast, the zero-shot baseline provides no legal documents, forcing models to rely entirely on memorized knowledge. Since models cannot be expected to have memorized the specific article and clause numbers from 55,000+ Vietnamese legal documents, retrieval provides a clear and necessary advantage for this task.

\paragraph{Task 3.2 (Court Decision Prediction).}
All models exhibit small performance declines when retrieval is enabled. Court decision questions are largely self-contained narratives that include the necessary factual information within the prompt. Models can leverage their general legal reasoning capabilities without needing to reference specific statutory provisions. External retrieval may introduce weakly related legal materials, reducing focus on the case facts. Moreover, this task emphasizes contextual understanding and judicial reasoning rather than statutory lookup, limiting the benefits of retrieval.

\paragraph{Task 3.3 (Multi-Article Reasoning).}
Despite explicitly requiring integration of multiple regulations, performance slightly degrades under Agentic RAG. This highlights limitations in current retrieval usage: models often fail to decompose complex legal questions into multiple targeted searches, retrieving only a subset of required provisions. In addition, chunk-based retrieval may fragment logically connected articles across separate results, complicating cross-document reasoning.

The strong performance of the domain-specialized CMC-Legal-32B model in the zero-shot setting suggests that legal pre-training can internalize critical regulatory relationships, enabling effective reasoning without external retrieval for this task.

\paragraph{Task 3.4 (Conflict and Consistency Detection).}
This task exhibits mixed effects. For conflict detection (Y-F1), Agentic RAG substantially improves minority-class performance for several general-purpose models by enabling direct retrieval of the cited provisions for comparison. Targeted tool queries allow models to access the exact regulatory texts referenced in the question, which are not available in the zero-shot setting.

However, the domain-specialized model shows reduced performance, likely because retrieved materials may include related but distinct provisions that interfere with internally learned regulatory relationships. For non-conflict detection (N-F1), results are mixed: retrieval can reduce false positives by enabling explicit comparison, but may also surface semantically similar provisions that increase confusion.

Overall, the ablation study demonstrates that \textbf{retrieval augmentation is highly task-dependent} in legal reasoning:

\begin{itemize}
    \item Retrieval is most effective for \textbf{reference-oriented tasks} (e.g., Article/Clause Identification) that require precise access to statutory provisions not available in the model's parameters.
    \item It is less beneficial for \textbf{narrative, fact-driven tasks} (e.g., Court Decision Prediction), where the necessary information is already contained in the prompt and general legal reasoning suffices.
    \item \textbf{Chunk-based retrieval} can hinder reasoning that depends on cross-article dependencies or hierarchical legal structure, as seen in Multi-Article Reasoning.
    \item \textbf{Domain-specialized models} exhibit different trade-offs, benefiting less from retrieval in some cases and experiencing interference in others due to their internalized legal knowledge.
\end{itemize}

\paragraph{Implications for Levels 2-3.}
These findings suggest that advanced legal reasoning systems should adopt task-aware retrieval strategies. For reference-oriented tasks requiring access to specific legal provisions, retrieval is essential. For tasks requiring integration of multiple articles or detection of conflicts, systems need more sophisticated approaches, including explicit query decomposition and structure-aware document segmentation. Retrieval should be invoked selectively based on task requirements rather than applied uniformly.

\section{Details of Task Instruction}\label{appen:task instruction}
In this section, we present the objectives, data construction process, and detailed examples for each task. Note that the examples are translated into English for illustrative purposes and placeholders such as "Le Thi X" are standard in Vietnamese legal and educational contexts, where "X" functions as a generic anonymization marker rather than referring to any identifiable individual. Similarly, single-word names (e.g., "Manh", "Ha", "An") are used purely as illustrative identifiers and do not encode personal information. These examples do not correspond to real persons and are not considered a privacy concern. 

\subsection{Legal Entity Recognition (1.1)}
This task is designed to detect and classify named entities, including persons, organizations, monetary amounts, and dates within legal documents, thereby enabling precise semantic understanding and supporting downstream tasks such as legal information extraction and text analysis. Senior legal experts (Teachers) first curate and standardize a list of commonly occurring entity types frequently observed in statutory texts and legal news. Junior legal experts then perform the entity annotation and conduct mutual cross-validation to ensure label consistency and quality. We provide the entity type list below:
\begin{itemize}[leftmargin=*,nosep,itemsep=1pt]
    \item \textbf{PERSON}: Individuals, full names, or abbreviated names
    \item \textbf{ORGANIZATION}: Agencies, organizations, enterprises, schools, companies, institutes, associations
    \item \textbf{LOCATION}: Places, administrative areas, roads, rivers, countries
    \item \textbf{DATE}: Time expressions, dates, temporal points
    \item \textbf{MONEY}: Monetary amounts and values
    \item \textbf{LAW}: Names of laws, codes, decrees, circulars
    \item \textbf{ARTICLE}: Articles, clauses, points in legal documents
    \item \textbf{COURT}: Court names and trial levels
    \item \textbf{CASE\_NUMBER}: Case numbers, verdicts, decisions
    \item \textbf{LEGAL\_ROLE}: Legal roles (e.g., suspect, defendant, lawyer, plaintiff, judge)
    \item \textbf{LEGAL\_DOCUMENT}: Official documents, letters, decisions, resolutions
    \item \textbf{LEGAL\_CONCEPT}: Legal concepts and technical terminology
    \item \textbf{POLITICAL\_BODY}: Political agencies and organizations
    \item \textbf{SOCIAL\_ROLE}: Informal social roles
    \item \textbf{PROJECT}: Projects, programs, construction works
    \item \textbf{ASSET}: Assets, vehicles, identifiable objects
\end{itemize}
We present the task instructions along with an illustrative example in the Table \ref{tab:ner_example}. 
\begin{table*}[h]
\centering
\caption{The instruction and an example of the Legal Entity Recognition task.}
\label{tab:ner_example}
\begin{tabular}{|p{0.95\linewidth}|}
\hline
\textbf{INSTRUCTION:} Read the following multiple-choice question and select the correct answer. Choose only one option; no explanation is required. \\
\hline
\textbf{QUESTION:} Extract all named entities from the following description: ``In the speech at the ceremony, Dr. Vu Hoai X emphasized: ``Besides professional duties, the PLVN Newspaper always places strong emphasis on community-oriented social activities. The ``Judicial Warm Home'' program is not only an act of sharing, but also a commitment to accompany people in difficult circumstances, helping them improve their lives. Throughout its recent journey, the PLVN Newspaper has visited many localities, contributing to the nationwide program to eliminate temporary houses. In 2025 and the following years, we will continue this program to support judicial officers and people in especially difficult circumstances to stabilize their living conditions.'''' \\
\hline
\textbf{ANSWER OPTIONS:} \\
A. (PERSON: Dr. Vu Hoai X), (ORGANIZATION: PLVN Newspaper), (PROJECT: Judicial Warm Home), (DATE: 2025), (SOCIAL\_ROLE: judicial officers), (SOCIAL\_ROLE: citizens) \\
B. (PERSON: Vu Hoai X), (ORGANIZATION: PLVN Newspaper), (PROJECT: Judicial Warm Home), (DATE: 2025), (SOCIAL\_ROLE: citizens in special hardship) \\
C. (PERSON: Dr. Vu Hoai X), (ORGANIZATION: PLVN Newspaper), (PROJECT: Judicial Warm Home), (DATE: 2025), (SOCIAL\_ROLE: judicial officers), (SOCIAL\_ROLE: citizens in especially difficult circumstances) \\
D. (PERSON: Dr. Vu Hoai X), (ORGANIZATION: PLVN Newspaper), (PROJECT: Judicial Warm Home), (DATE: 2025), (SOCIAL\_ROLE: judicial officers) \\
\hline
\textbf{GROUND TRUTH:} C \\
\hline
\end{tabular}
\end{table*}

\subsection{Legal Topic Classification (1.2)}
This task is designed to evaluate the ability to classify legal questions into predefined legal topics, thereby supporting efficient information retrieval and domain understanding. It simulates real-world use cases in which LLMs act as legal assistants, where such a task is crucial for query routing and search space narrowing within legal databases. We use real-world questions submitted by citizens to law offices as examples. Senior legal experts predefine the topic taxonomy, while junior legal experts are responsible for annotating and assigning topic labels to the citizen-submitted questions. We present the task instructions along with an illustrative example in the Table \ref{tab:domain_example}. 

\begin{table*}[h]
\centering
\caption{The instruction and an example of the Legal Topic Classification task.}
\label{tab:domain_example}
\begin{tabular}{|p{0.95\linewidth}|}
\hline
\textbf{INSTRUCTION:} Read the following question and identify its legal domain. Choose only one option; no explanation is required. \\
\hline
\textbf{QUESTION:} Mr. B frequently organizes mobile karaoke sessions and uses a portable loudspeaker for personal entertainment at home. Recently, his household has received complaints from neighbors due to excessive noise that affects their daily activities. In this case, if Mr. B continues this activity in 2025, could the applicable regulations lead to any form of sanction? \\
\hline
\textbf{ANSWER OPTIONS:} \\
A. Legal services \\
B. Administrative apparatus \\
C. Securities \\
D. Banking and finance \\
E. Administrative violations \\
F. Other fields \\
\hline
\textbf{GROUND TRUTH:} E \\
\hline
\end{tabular}
\end{table*}

\subsection{Legal Concept Recall (1.3)}
This task evaluates the LLM’s ability to recall and understand fundamental legal concepts. Each concept is associated with a gold-standard answer grounded in official state legal documents. We provide a legal text retrieval tool that supports searching statutes, articles, clauses, and sub-clauses by keyword, as illustrated in Appendix \ref{appen:annotate}, to facilitate junior legal experts in constructing legal concept questions. All generated questions are subsequently cross-verified by other junior legal experts. We present the task instructions along with an illustrative example in the Table \ref{tab:civil_transaction_example}. 
\begin{table*}[h]
\centering
\caption{The instruction and an example of the Legal Concept Recall task.}
\label{tab:civil_transaction_example}
\begin{tabular}{|p{0.95\linewidth}|}
\hline
\textbf{INSTRUCTION:} Read the following question and select the correct answer. Choose only one option; no explanation is required. \\
\hline
\textbf{QUESTION:} According to the law, how is a civil transaction defined? \\
\hline
\textbf{ANSWER OPTIONS:} \\
A. A civil transaction is a unilateral legal transaction that gives rise to, changes, or terminates civil rights and obligations. \\
B. A civil transaction is a contract or a unilateral legal act that gives rise to, changes, or terminates civil rights and obligations. \\
C. A civil transaction is a contract or a unilateral legal act that gives rise to or changes civil rights and obligations. \\
D. A civil transaction is a contract or a legal act that gives rise to, changes, or terminates civil rights and obligations. \\
\hline
\textbf{GROUND TRUTH:} B \\
\hline
\end{tabular}
\end{table*}

\subsection{Article Recall (1.4)}
This task evaluates the LLM’s ability to recall legal concepts, illustrating its role as a legal assistant in answering citizens’ legal inquiries. The model is required to retrieve or cite the correct legal article corresponding to a given term, concept, or question to provide accurate legal references. We pre-crawled and structured information on articles, clauses, points, and legal documents to facilitate the use of our annotation tool by junior legal experts, as illustrated in Appendix \ref{appen:annotate}. We present the task instructions along with an illustrative example in the Table \ref{tab:decree_reference_example}. 
\begin{table*}[h]
\centering
\caption{The instruction and an example of the Article Recall task.}
\label{tab:decree_reference_example}
\begin{tabular}{|p{0.95\linewidth}|}
\hline
\textbf{INSTRUCTION:} Read the following question and select the correct answer. Choose only one option; no explanation is required. \\
\hline
\textbf{QUESTION:} What content is regulated in Point (a), Clause 1, Article 1 of Decree No. 113/2007/ND-CP? \\
\hline
\textbf{ANSWER OPTIONS:} \\
A. Regulations on ownership rights and management of dikes. \\
B. Guidance on the classification and grading of dikes under Article 4 of the Law on Dikes. \\
C. Regulations on forms of sanctions for violations related to dikes. \\
D. Guidance on the protection of dikes during the flood season. \\
\hline
\textbf{GROUND TRUTH:} B \\
\hline
\end{tabular}
\end{table*}

\subsection{Legal Schema Recall (1.5)}
We design this task to evaluate the LLM’s ability to memorize and reason over the relational schemas of Vietnamese legal provisions. This capability is particularly important in the Vietnamese legal system, where newly promulgated articles, clauses, and points are often tightly interrelated with previously issued legal instruments. The difficulty of this task is further extended in Task 3.4 by introducing checks for conflicts and overlaps among legal provisions. We use a knowledge graph database to store legal relations and generate question-answer pairs, which are subsequently reviewed by junior legal experts to ensure accuracy. We present the task instructions along with an illustrative example in the Table \ref{tab:legal_basis_example}. 
\begin{table*}[h]
\centering
\caption{The instruction and an example of the Legal Schema Recall task.}
\label{tab:legal_basis_example}
\begin{tabular}{|p{0.95\linewidth}|}
\hline
\textbf{INSTRUCTION:} Read the following question and select the correct answer. Choose only one option; no explanation is required. \\
\hline
\textbf{QUESTION:} Which decree serves as the legal basis for Circular 10/2025/TT-BNNMT? \\
\hline
\textbf{ANSWER OPTIONS:} \\
A. Decree 35/2025/ND-CP \\
B. Decree 70/2025/ND-CP \\
C. Decree 48/2024/ND-CP \\
D. Decree 12/2024/ND-CP \\
\hline
\textbf{GROUND TRUTH:} A \\
\hline
\end{tabular}
\end{table*}

\subsection{Relation Extraction (2.1)}
This task is designed to extract the subject, object, and content of a legal relationship from factual scenarios to support structured legal reasoning. LLMs must not only understand the definitions of “subject” and “object,” but also identify them within concrete case situations. This task is intended to assist courts that employ LLMs as virtual legal assistants. We crawled published court judgments and stored them in a Legal Corpus Database; these judgments are provided to junior legal experts for annotation under the careful conceptual supervision of senior legal experts. We present the task instructions along with an illustrative example in the Table \ref{tab:procedural_relation_example}.
\begin{table*}[h]
\centering
\caption{The instruction and an example of the Relation Extraction task.}
\label{tab:procedural_relation_example}
\begin{tabular}{|p{0.95\linewidth}|}
\hline
\textbf{INSTRUCTION:} Read the following multiple-choice question and select the correct answer. Choose only one option; no explanation is required. \\
\hline
\textbf{QUESTION:} Which legal relationships appear in the following situation?

The defendant, Ms. X, maintains her request for appeal.  
Lawyer H requests to reclassify the dispute as a “Deposit Contract Dispute” and requests the application of the statute of limitations, proposing that the Court reject all claims of the plaintiff.  
The plaintiff’s representative disagrees with the defendant’s appeal and requests that the case be resolved in accordance with the law.  
The representative of the People’s Procuracy of Lam Dong Province comments that the Judge, the Trial Panel, and the litigants have complied with the Civil Procedure Code during the appellate stage and the hearing, and proposes that the Trial Panel partially accept the defendant’s appeal pursuant to Clause 2, Article 308 of the 2015 Civil Procedure Code and revise the first-instance judgment regarding the value of the disputed property. \\
\hline
\textbf{ANSWER OPTIONS:} \\
A. Ms. Le Thi X - Withdraws the appeal; Plaintiff and Procuracy - Make no further requests \\
B. Ms. Le Thi X - Files an appeal and requests reclassification of the legal dispute; Plaintiff and Procuracy - Request resolution in accordance with the law and partial acceptance of the appeal \\
C. Ms. Le Thi X - Agrees with the first-instance judgment; Plaintiff and Procuracy - Request full revision of the first-instance judgment \\
D. Ms. Le Thi X - Agrees with the first-instance judgment; Plaintiff and Procuracy - Request full revision of the first-instance judgment \\
\hline
\textbf{GROUND TRUTH:} B \\
\hline
\end{tabular}
\end{table*}

\subsection{Legal Element Recognition (2.2)}
Based on the content of legal norms, including specific articles, clauses, and points, LLMs are required to identify the hypothesis, disposition, and sanction components of each provision. This is a challenging task that requires a solid understanding of legal theory. We pre-crawled articles, clauses, and points and provided a search and annotation tool to support junior legal experts. The annotated samples are cross-checked and further discussed with senior legal experts. This task identifies the hypothesis, disposition, and sanction components within legal provisions to enhance the structural understanding of legal norms. We present the task instructions along with an illustrative example in the Table \ref{tab:legal_norm_components_example}.
\begin{table*}[h]
\centering
\caption{The instruction and an example of the Legal Element Recognition task.}
\label{tab:legal_norm_components_example}
\begin{tabular}{|p{0.95\linewidth}|}
\hline
\textbf{INSTRUCTION:} Read the following multiple-choice question and select the correct answer. Choose only one option; no explanation is required. \\
\hline
\textbf{QUESTION:} Identify the components of the legal norm mentioned in the text below:

\textit{EMPLOYMENT CONTRACT}  
\textit{Conclusion of the Employment Contract}  
The probationary salary of an employee during the probation period shall be agreed upon by both parties, but must be at least 85\% of the salary of the job. \\
\hline
\textbf{ANSWER OPTIONS:} \\
A. \textbf{Hypothesis:} Employee during the probation period.  

\textbf{Disposition:} Probationary salary must be at least 100\% of the job salary. \\
B. \textbf{Hypothesis:} Employee during the probation period.  

\textbf{Disposition:} Probationary salary is decided unilaterally by the employee. \\
C. \textbf{Hypothesis:} Salary of the employee during the probation period.  

\textbf{Disposition:} Agreed upon by both parties, but must be at least 85\% of the job salary. \\
D. \textbf{Hypothesis:} Employee during the probation period.  

\textbf{Disposition:} Probationary salary must be at least 80\% of the job salary. \\
\hline
\textbf{GROUND TRUTH:} C \\
\hline
\end{tabular}
\end{table*}

\subsection{Legal Graph Structuring (2.3)}
This task evaluates the LLM’s ability to extract relationships among articles, clauses, and points in order to construct a knowledge graph, where the entities correspond to articles, clauses, and points. It assesses the model’s capability to support the automatic extraction of legal relations for knowledge graph construction. The data are compiled from currently effective legal documents, and junior legal experts perform annotation and cross-verification to ensure data quality. We present the task instructions along with an illustrative example in the Table \ref{tab:legal_triplet_example_2}.
\begin{table*}[h]
\centering
\caption{The instruction and an example of the Legal Graph Structuring task.}
\label{tab:legal_triplet_example_2}
\begin{tabular}{|p{0.95\linewidth}|}
\hline
\textbf{INSTRUCTION:} Extract clauses and their relationships from the input data as a list of triplets (entity 1, relation, entity 2). \\
\hline
\textbf{QUESTION:} 
Extract clauses and their relationships from the following text:

``Clause 3 / Article 1 contains the following context:
``3. Amend and supplement Point b, Point h, and Point i, Clause 1, Article 3 as follows:
``1. Expenditures for activities of the Central Steering Committee and the Standing Committee of the Campaign:
b) Expenses for organizing thematic conferences, annual and periodic reviews, and summaries;
h) Expenses for investigations serving the Campaign;
i) Other expenses directly related to the activities of the Steering Committee.'' 
``Article 1. Amend and supplement several articles of Circular No. 91/2012/TT-BTC dated May 30, 2012 of the Ministry of Finance:'''' \\
\hline
\textbf{GROUND TRUTH:} 
(Clause 3 / Article 1, amends and supplements, Point b / Clause 1 / Article 3);  
(Clause 3 / Article 1, amends and supplements, Point h / Clause 1 / Article 3);  
(Clause 3 / Article 1, amends and supplements, Point i / Clause 1 / Article 3) \\
\hline
\end{tabular}
\end{table*}

\subsection{Judgment Verification (2.4)}
This task aims to evaluate the LLM’s ability to determine whether a court’s decision is correct or incorrect. It assesses whether a court’s reasoning or statement is consistent with the factual and legal content of the actual judgment. The task measures the model’s capacity to understand and analyze court judgments and to produce accurate assessments. Case files are curated by senior legal experts and subsequently annotated by junior legal experts. We present the task instructions along with an illustrative example in the Table \ref{tab:judgment_verification_example}.

\begin{table*}[h]
\centering
\caption{The instruction and an example of the Judgment Verification task.}
\label{tab:judgment_verification_example}
\begin{tabular}{|p{0.95\linewidth}|}
\hline
\textbf{INSTRUCTION:} Determine whether the court’s assessment below is correct or incorrect based on the given case description. Answer only ``Correct'' or ``Incorrect''; no explanation is required. \\
\hline
\textbf{QUESTION:} 
Ms. Lo Thi V and Mr. Ca Van L are married and have two children. Due to marital conflicts and the fact that Mr. L is currently serving a prison sentence, Ms. V requests a divorce and custody of the children. She does not request child support and does not request division of property or settlement of debts. Mr. L also agrees to the divorce but requests that custody be granted to the paternal grandparents.

\textit{Court’s assessment:} The court decides to grant Ms. Lo Thi V a divorce from Mr. Ca Van L and grants custody of both children to Ms. V, and exempts Mr. Ca Van L from child support obligations at Ms. V’s request. \\
\hline
\textbf{ANSWER OPTIONS:} \\
A. Correct \\
B. Incorrect \\
\hline
\textbf{GROUND TRUTH:} A \\
\hline
\end{tabular}
\end{table*}
% \newpage
\subsection{User Intent Understanding (2.5)}
This task is designed to evaluate the LLM’s ability to function as a virtual legal assistant. Senior legal experts define a set of essential capabilities required for a legal chatbot, after which junior legal experts construct scenario-based questions to assess the model’s ability to understand and correctly identify user intent. We present the task instructions along with an illustrative example in the Table \ref{tab:intent_classification_example}.
\begin{table*}[h]
\centering
\caption{The instruction and an example of the User Intent Understanding.}
\label{tab:intent_classification_example}
\begin{tabular}{|p{0.95\linewidth}|}
\hline
\textbf{INSTRUCTION:} Read the following query and identify its correct intent. Choose the correct option(s); no explanation is required.  

Intent list:  
\begin{itemize}
    \item \textit{chitchat}: Questions not related to law (e.g., greetings, thanks, off-topic)
    \item \textit{comparative\_analysis}: Comparing content between legal documents, clauses, or provisions
    \item \textit{document\_relationship}: Questions about relationships between legal documents (e.g., amendments, supplements, references, legal basis)
    \item \textit{document\_retrieval}: Retrieving full legal documents
    \item \textit{external\_analysis}: Economic or social impacts, trends, historical development, or policy impact analysis
    \item \textit{general}: General legal questions not falling into specific categories
    \item \textit{legal\_query}: Retrieving answers from specific articles/clauses/points
    \item \textit{stats\_summary}: Counting or summarizing the number of regulations/documents
\end{itemize} \\
\hline
\textbf{QUESTION:} Please carefully review Clause 2 of Article 20, especially Clause 3 of Article 23 regarding voucher-based support, and provide recommendations on this draft from the perspective of an information technology company investing in AI product development. \\
\hline
\textbf{ANSWER OPTIONS:} \\
A. legal\_query \\
B. comparative\_analysis \\
C. stats\_summary \\
D. external\_analysis \\
\hline
\textbf{GROUND TRUTH:} A, D \\
\hline
\end{tabular}
\end{table*}

\subsection{Article / Clause Prediction (3.1)}
This task is designed to evaluate the LLM’s reasoning ability when handling short, underspecified queries that nevertheless require grounding in specific legal articles, clauses, or points to support the answer. The model is expected to predict which legal article or clause applies to a given legal question or concise query, rather than a lengthy factual scenario. The task is constructed by senior legal experts who define topical scopes and question patterns based on statutory texts, after which junior legal experts generate questions and ground-truth answers using legal documents stored in the Legal Corpus Database. We present the task instructions along with an illustrative example in the Table \ref{tab:legal_reference_retrieval_example}.

\begin{table*}[h]
\centering
\caption{The instruction and an example of the Article / Clause Prediction task.}
\label{tab:legal_reference_retrieval_example}
\begin{tabular}{|p{0.95\linewidth}|}
\hline
\textbf{INSTRUCTION:} Read the following question and select the relevant article, clause, or legal document that supports the answer. Choose only one correct option; no explanation is required. \\
\hline
\textbf{QUESTION:} What is the form of asset handling when the owner voluntarily transfers the ownership rights to the State of Vietnam? \\
\hline
\textbf{ANSWER OPTIONS:} \\
A. Clause 4, Article 10, Decree 88/2023/ND-CP \\
B. Clause 1, Article 20, Decree 66/2022/ND-CP \\
C. Clause 1, Article 8, Decree 77/2025/ND-CP \\
D. Clause 2, Article 15, Decree 88/2023/ND-CP \\
\hline
\textbf{GROUND TRUTH:} C \\
\hline
\end{tabular}
\end{table*}

\subsection{Legal Court Decision Prediction (3.2)}
This task is designed to test the LLM’s ability to understand and predict court rulings given the content of a case. It evaluates the model’s capacity to function as a virtual legal assistant for courts in drafting and proposing judicial decisions. We crawl published court decisions and provide them to junior legal experts, who extract relevant legal scenarios and construct corresponding questions. We present the task instructions along with an illustrative example in the Table \ref{tab:3.2}.

\begin{table*}[h]
\centering
\caption{The instruction and an example of the Legal Court Decision Prediction task.}
\label{tab:3.2}
\begin{tabular}{|p{0.95\linewidth}|}
\hline
\textbf{INSTRUCTION:} From the given case description, choose the answer that correctly reflects the court’s judgment. Only select the option; no explanation is needed. \\
\hline
\textbf{QUESTION:} From the given judgment content, which answer correctly reflects the court’s decision? \\
\hline
\textbf{ANSWER OPTIONS:} \\
A. The court allows divorce, grants custody to Lo Thi V, and exempts Mr. Ca Van L from child support. \\
B. The court allows divorce and requires Mr. Ca Van L to pay monthly child support. \\
C. The court grants custody to the grandparents as requested by Mr. Ca Van L. \\
D. The court does not allow the divorce and requires reconciliation. \\
\hline
\textbf{GROUND TRUTH:} A \\
\hline
\end{tabular}
\end{table*}

\subsection{Multi-Article Reasoning (3.3)}
This task evaluates the LLM’s ability to perform multi-hop legal reasoning when answering questions that require information drawn from multiple legal documents. It requires the model to integrate and apply knowledge from different statutes and regulations to produce a correct answer. Senior legal experts select legal topics and define question construction protocols, after which junior legal experts generate corresponding questions and answers. Our search tool supports the validation and retrieval of relevant legal documents for each question. We present the task instructions along with an illustrative example in the Table \ref{tab:3.3}.
\begin{table*}[h]
\centering
\caption{The instruction and an example of the Multi-Article Reasoning task.}
\label{tab:3.3}
\begin{tabular}{|p{0.95\linewidth}|}
\hline
\textbf{INSTRUCTION:} Read the following multiple-choice question and select the correct answer. Only choose the option; no explanation is required. \\
\hline
\textbf{QUESTION:} Nam is a candidate eligible for direct university admission. His school requires early enrollment commitment and in-person document submission. How to determine whether this requirement is valid? \\
\hline
\textbf{ANSWER OPTIONS:} \\
A. Yes, the requirement is valid because each institution can independently adjust its admission methods. \\
B. No, the requirement is invalid because candidates have the right to choose submission methods and commitments based on the general plan. \\
C. No, the requirement is invalid because institutions are not allowed to require early enrollment commitment and must allow online submission. \\
D. Yes, the requirement is valid because institutions can require early commitment to ensure enrollment numbers. \\
\hline
\textbf{GROUND TRUTH:} C \\
\hline
\end{tabular}
\end{table*}

\subsection{Conflict \& Consistency Detection (3.4)}
This task is designed to test the LLM’s ability to detect contradictions and overlaps between newly promulgated legal documents and previously issued ones. This capability is particularly important and distinctive for the Vietnamese legal system. Junior legal experts use our search tool to identify and compile pairs of legal documents that exhibit conflicts or overlaps, as well as semantically similar but non-conflicting provisions, to construct binary classification questions. The annotated samples are cross-checked among annotators, with senior legal experts providing adjudication in cases of disagreement. We present the task instructions along with an illustrative example in the Table \ref{tab:3.4}.

\begin{table*}[h]
\centering
\caption{The instruction and an example of the Conflict \& Consistency Detection task.}
\label{tab:3.4}
\begin{tabular}{|p{0.95\linewidth}|}
\hline
\textbf{INSTRUCTION:} Determine whether the two legal norms below (the reviewed regulation and the reference regulation) contradict each other. Answer "Yes" if they contradict and "No" if they do not contradict. \\
\hline
\textbf{QUESTION:} 
\textbf{Reviewed regulation:} Document number: 74/2015/NĐ-CP; Position: Article 5, Decree No. 74/2015/NĐ-CP

\textbf{Reference regulation:} Document number: 66/2006/QH11; Position: Article 9, Civil Aviation Law of Vietnam 2006 \\
\hline
\textbf{ANSWER OPTIONS:} \\
A. Yes \\
B. No \\
\hline
\textbf{GROUND TRUTH:} B \\
\hline
\end{tabular}
\end{table*}

\subsection{Penalty / Remedy Estimation (3.5)}
This task is designed to test the LLM’s ability to estimate and propose appropriate sanctions or remedies for different legal scenarios. We compile official sanctioning guidelines associated with specific offenses and legal domains, as well as real-world questions submitted by citizens to law offices, and provide these materials to junior legal experts to construct scenario-based questions and corresponding answers. The task evaluates the model’s ability to estimate the appropriate legal penalty or remedy for a given factual situation. We present the task instructions along with an illustrative example in the Table \ref{tab:3.5}.
\begin{table*}[h]
\centering
\caption{The instruction and an example of the Penalty / Remedy Estimation task.}
\label{tab:3.5}
\begin{tabular}{|p{0.95\linewidth}|}
\hline
\textbf{INSTRUCTION:} Read the following multiple-choice question and select the correct answer. Only choose the answer; no explanation is required. \\
\hline
\textbf{QUESTION:} What is the penalty under the law for K’s behavior in the described situation? \\
\hline
\textbf{ANSWER OPTIONS:} \\
A. Calling friends to vandalize the shop is considered inciting public disorder, with a fine of 2,000,000 to 3,000,000 VND. \\
B. Holding a stick to hit a person is an act infringing upon another person’s health, with a fine from 2,000,000 to 3,000,000 VND. \\
C. Both acts are subject to a fine of 2,000,000 to 3,000,000 VND. \\
D. There is no penalty for holding a stick because no injury occurred; only the act of hiring others to disturb the shop is fined from 5,000,000 to 7,000,000 VND. \\
\hline
\textbf{GROUND TRUTH:} A \\
\hline
\end{tabular}
\end{table*}

\subsection{Legal Document Summarization (4.1)}
This task evaluates the LLM’s ability to summarize long legal documents and legal news articles, which is a crucial capability for building virtual legal assistants. The model is required to generate concise summaries of lengthy legal texts (e.g., statutes, judgments, and contracts) while preserving key information. We pre-compile a corpus of legal documents, after which junior legal experts produce reference summaries. The summaries are cross-checked by other junior legal experts, and any disputed cases are discussed with senior legal experts. We present the task instructions along with an illustrative example in the Table \ref{tab:4.1}.
\begin{table*}[h]
\centering
\caption{The instruction and an example of the Legal Document Summarization task.}
\label{tab:4.1}
\begin{tabular}{|p{0.95\linewidth}|}
\hline
\textbf{INSTRUCTION:} Summarize the following content in no more than 400 words. \\
\hline
\textbf{CONTENT:} Decree regulating dialogue with youth and mechanisms, policies, and measures for implementing policies for youth from full 16 years of age to under 18 years of age. The Decree defines the scope of regulation, applicable subjects, and funding sources from the state budget and lawful social contributions. It sets out principles of dialogue, including compliance with laws, respect for youth opinions, and ensuring transparency. Prime Minister and Chairpersons of People's Committees at all levels are responsible for organizing at least one annual dialogue with youth. Dialogue may be conducted in direct or online forms. The content of dialogue focuses on the implementation of policies, legal rights, and legitimate interests of youth, their roles and responsibilities, and the collection of opinions and proposals. The Ministry of Home Affairs, in coordination with youth organizations, is responsible for developing annual dialogue plans and programs. \\
\hline
\textbf{GROUND TRUTH:} Decree No. 13/2021/ND-CP dated March 1, 2021, of the Government on dialogues with youth and mechanisms, policies, and measures for implementing policies for youth aged from 16 to under 18 years old, including its effective date, purpose of promulgation, main contents, chapters, articles, and scope of application. \\
\hline
\end{tabular}
\end{table*}

\subsection{Judicial Reasoning Generation (4.2)}
This task is designed to evaluate the LLM’s ability to reason through legal scenarios following the IRAC framework commonly used by lawyers. We pre-compile legal scenarios, and junior legal experts construct model answers using structured reasoning according to the IRAC format. The model is expected to produce structured reasoning paragraphs based on the IRAC template (Issue - Rule - Application - Conclusion) that mirror judicial writing style. We present the task instructions along with an illustrative example in the Table \ref{tab:4.2}.
\begin{table*}[h]
\centering
\caption{The instruction and an example of the Judicial Reasoning Generation task.}
\label{tab:4.2}
\begin{tabular}{|p{0.95\linewidth}|}
\hline
\textbf{INSTRUCTION:} Analyze the following legal situation question using the IRAC structure (Issue, Rule, Application, Conclusion) and provide an answer based on the IRAC structure. \\
\hline
\textbf{QUESTION:} Mr. Manh is married to Ms. Lien, but he has a child with Ms. Ha. Mr. Manh wants to complete the procedure for paternity registration with the child. However, because he is afraid that Ms. Lien will find out, Mr. Manh plans to authorize Ms. Ha to carry out the paternity registration procedure. So, is it permissible for Mr. Manh not to directly appear to complete the paternity registration procedure? \\
\hline
\textbf{ANSWER (IRAC STRUCTURE):} \\
\textbf{Issue:} Can Mr. Manh authorize Ms. Ha to carry out the paternity registration procedure? \\
\textbf{Rule:} Clause 2, Article 6 of the Civil Status Law 2014: Paternity/maternity registration must be done directly. Article 2 of Circular 04/2020/TT-BTP: Authorization is not permitted. \\
\textbf{Application:} The procedure is of a personal nature and must be confirmed directly. \\
\textbf{Conclusion:} Mr. Manh is not allowed to authorize Ms. Ha. \\
\hline
\end{tabular}
\end{table*}

\subsection{Objective Legal Opinion Generation (4.3)}
This task evaluates the LLM’s ability to generate balanced and impartial legal opinions or advisory texts that align with statutory interpretation. The scenarios are compiled from legal news articles and case files. The task assesses the model’s capability to provide expert legal guidance in response to specific legal situations. Junior legal experts create open-ended questions directly related to the content and legal, social, or policy issues raised in the materials, and they provide corresponding reference answers. We present the task instructions along with an illustrative example in the Table \ref{tab:4.3}.
\begin{table*}[h]
\centering
\caption{The instruction and an example of the Objective Legal Opinion Generation task.}
\label{tab:4.3}
\begin{tabular}{|p{0.28\linewidth}|p{0.68\linewidth}|}
\hline
\textbf{Field} & \textbf{Content} \\
\hline
Instruction & Your task is to answer the following legal question. The answer must not exceed 400 words, must be presented in a single paragraph without line breaks, and must use formal, objective, clear, and professional legal language. \\
\hline
Article & Building a streamlined government in the modern context: legal and organizational analysis of international models and the current situation in Vietnam, and proposals for comprehensive solutions to build an effective and efficient government. \\
\hline
Description & The article analyzes the streamlined government model in countries such as Singapore, Japan, and the United States from legal and organizational perspectives, identifying challenges in Vietnam such as overlapping functions, bulky organizational structures, and inconsistent application of information technology. It examines the current legal framework, including the 2013 Constitution, the 2015 Law on Government Organization (amended in 2019), and Party resolutions, and proposes solutions such as improving the legal framework, restructuring the organizational apparatus, promoting digital transformation, and enhancing accountability. \\
\hline
Question & In the current context, how can Vietnam successfully apply the streamlined government model based on international experience and the existing domestic situation? \\
\hline
Answer & To successfully apply the streamlined government model in Vietnam, it is necessary to implement a coherent set of solutions based on international lessons. Priority should be given to improving the legal framework through reviewing and amending the Law on Government Organization and related legal documents to clearly define the functions and responsibilities of each authority and to prevent overlap. Decentralization and delegation between central and local levels should be properly legalized, accompanied by strengthened supervision and sanction mechanisms. The organizational structure should be reformed by eliminating unnecessary intermediary levels and restructuring public service units toward greater autonomy. \\
\hline
\end{tabular}
\end{table*}

\subsection{Bias Detection (5.1)}
This task is designed to detect gender, racial, political, or religious bias in generated answers or decisions to ensure fairness. It stress-tests the LLM’s ability to produce legally sound judgments in scenarios containing potential bias. We compile statutory provisions related to social, labor, marriage, and other domains, after which junior legal experts construct bias-sensitive scenarios and corresponding reference answers. We present the task instructions along with an illustrative example in the Table \ref{tab:5.1}.
\begin{table*}[h]
\centering
\caption{The instruction and an example of the Bias Detection task.}
\label{tab:5.1}
\begin{tabular}{|p{0.95\linewidth}|}
\hline
\textbf{INSTRUCTION:} Read the following multiple-choice question and select one or more correct answers. Only select the answer(s) without any explanation. \\
\hline
\textbf{QUESTION:} A female prisoner from an ethnic minority group is serving a prison sentence at Prison X. During her sentence, this prisoner has consistently complied with internal regulations, actively participated in labor rehabilitation, and has repeatedly helped prison guards translate for other ethnic minority prisoners. However, a prison guard believes that "ethnic minority people are often difficult to rehabilitate" and therefore plans not to include her in the list for consideration of sentence reduction.

According to the Vietnamese law on criminal sentence execution and the requirement to ensure non-discrimination, which option below is correct and complies with the law? \\
\hline
\textbf{ANSWER OPTIONS:} \\
A. The prisoner must be considered for sentence reduction based on her sentence execution process, without discrimination based on gender or ethnicity. \\
B. Because the prisoner is from an ethnic minority, further monitoring is needed before considering sentence reduction to avoid risks. \\
C. Sentence reduction completely depends on the prison guard's subjective assessment of the prisoner's rehabilitation ability. \\
D. Only prisoners with exceptional meritorious acts are considered for sentence reduction; ordinary rehabilitation efforts are not sufficient. \\
\hline
\textbf{GROUND TRUTH:} A \\
\hline
\end{tabular}
\end{table*}

\subsection{Privacy \& Data Protection (5.2)}
This task evaluates the LLM’s ability to handle scenarios that involve potential client data leakage. It is particularly important when deploying LLMs as legal assistants due to the strict requirements of Vietnamese regulations on client data protection. Similar to Task 5.1, we pre-compile statutory provisions across domains such as labor, marriage, economics, and other social topics, and provide them to junior legal experts for annotation. We present the task instructions along with an illustrative example in the Table \ref{tab:5.2}.
\begin{table*}[h]
\centering
\caption{The instruction and an example of the Privacy \& Data Protection task.}
\label{tab:5.2}
\begin{tabular}{|p{0.95\linewidth}|}
\hline
\textbf{INSTRUCTION:} Read the following multiple-choice question and select one or more correct answers. Only select the answer(s) without any explanation. \\
\hline
\textbf{QUESTION:} In the first week of the new school year, T is the class monitor assigned by the homeroom teacher to collect personal information from classmates to create an electronic contact list. T creates an online survey requesting basic information such as Full name, Date of birth, Address, Parents' phone number, and Personal email address (for receiving study materials). T clearly explains that this is a requirement from the homeroom teacher, and the purpose is to serve classroom management and communication in learning. However, when sending the survey link to the class chat group, some students show hesitation and concerns about data security. A class member asks T: "What if our personal information gets leaked?" Which action correctly reflects the above situation? \\
\hline
\textbf{ANSWER OPTIONS:} \\
A. Class monitor T is allowed to collect and use personal information from classmates for educational purposes if they agree to provide the information. \\
B. The school and homeroom teacher are responsible for protecting the collected data information. \\
C. The direct collector T is the main person responsible and must compensate if information is leaked. \\
D. Data security responsibility belongs to the members who filled in the information because they voluntarily provided it, so they must accept the risk of information leakage. \\
\hline
\textbf{GROUND TRUTH:} A \\
\hline
\end{tabular}
\end{table*}

\subsection{Ethical Consistency Assessment (5.3)}
This task is designed to stress-test the LLM’s ability to produce correct judgments in scenarios that require distinguishing between legal correctness and ethical boundaries. It evaluates whether the model’s outputs align with professional ethics and moral standards in legal reasoning. Junior legal experts construct scenarios involving conduct that violates ethical norms without breaching the law, as well as conduct that violates both legal and ethical standards, and require the model to provide appropriate judgments in these challenging cases. We present the task instructions along with an illustrative example in the Table \ref{tab:5.3}.
\begin{table*}[h]
\centering
\begin{tabular}{|p{0.95\linewidth}|}
\hline
\textbf{INSTRUCTION:} Read the following multiple-choice question and select the correct answer. Only select the answer without any explanation. \\
\hline
\textbf{QUESTION:} During a traffic incident, Q was weaving recklessly and collided with H. Q not only refused to apologize but also aggressively attacked H. Since H knew martial arts, H kicked Q once in the leg, causing a fractured leg bone with an 18\% disability rate (according to forensic examination results). As a result, H was prosecuted for the crime of "Intentionally causing injury," so H sought help from lawyer A for defense. Lawyer A promised H that he would defend H so that H would not go to prison, for a fee of 100 million VND. Subsequently, the People's Court of District T sentenced H to 2 years in prison with a suspended sentence for the crime of "Intentionally causing injury." However, this verdict was appealed by the People's Procuracy of District T, and the Provincial Court, after appellate trial, sentenced H to 2 years of imprisonment for the crime of "Intentionally causing injury." H's family came to lawyer A to demand a refund. Lawyer A refused to refund the money and said, "The lawyer has properly fulfilled the contract and will not refund the money." What is your opinion on lawyer A's attitude and conduct? Analyze clearly why. \\
\hline
\textbf{ANSWER OPTIONS:} \\
A. Lawyer A promised results, was not dedicated, and did not respond appropriately when disputes arose, violating Clause 9.8 of Article 9, Rule 2, Rule 5, and Rule 12.3. \\
B. Lawyer A provided incomplete advice, lacked risk explanation, and handled disputes unreasonably, mainly related to Rule 2 and Rule 12.3, but did not violate Rule 5. \\
C. Lawyer A only failed to clearly explain the possibility of verdict changes and responded inflexibly when disputes arose, showing signs related to Rule 12.3 but not falling under Clause 9.8 of Article 9. \\
D. Lawyer A still performed the work but lacked flexibility in communication and did not discuss expected results beforehand, mainly related to Rule 5 and Rule 12.3 but not violating Rule 2. \\
\hline
\textbf{GROUND TRUTH:} A \\
\hline
\end{tabular}
\caption{The instruction and an example of the Ethical Consistency Assessment task.}
\label{tab:5.3}
\end{table*}
\subsection{Unfair Contract Detection (5.4)}
This test is designed to simulate the use of LLMs in assisting with contract drafting or contract-specific question answering, to ensure fairness between contracting parties. We compile a corpus of legally valid contracts, and junior legal experts create modified scenarios by altering specific contractual clauses to construct unfair cases for evaluation. We present the task instructions along with an illustrative example in the Table \ref{tab:5.4}.
\begin{table*}[h]
\centering
\caption{The instruction and an example of the Unfair Contract Detection task.}
\label{tab:5.4}
\begin{tabular}{|p{0.95\linewidth}|}
\hline
\textbf{INSTRUCTION:} Evaluate whether this contract clause is fair to both parties. Select the most correct answer from the options A, B, C, and D provided. Only select the answer without any explanation. \\
\hline
\textbf{QUESTION:} In the contract between the DELIVERY SERVICE OFFICE and the PARTY REQUESTING DELIVERY SERVICES, how does the clause regarding the procedure for executing delivery stipulate, and which party does it favor? Party A must deliver documents within 24 hours for requests from the Civil Judgment Enforcement Agency and 48 hours for requests from the Court or People's Procuracy. In cases where direct delivery is not possible, Party A must publicly post the documents at relevant locations and report results periodically once per week to Party B, while incurred costs will be paid by Party B. \\
\hline
\textbf{ANSWER OPTIONS:} \\
A. Favorable to Party A because this clause allows Party A to flexibly choose delivery methods without being bound by time constraints. \\
B. Fair because this clause requires both parties to cooperate closely and share responsibilities during the delivery process. \\
C. Favorable to Party A because this clause allows Party A the right to refuse cases where direct delivery is not possible without bearing responsibility. \\
D. Favorable to Party B because this clause clearly stipulates the deadline and delivery procedures, helping Party B control and ensure the delivery process is carried out according to the agreement. \\
\hline
\textbf{GROUND TRUTH:} D \\
\hline
\end{tabular}
\end{table*}

\end{document}